\documentclass[conference]{IEEEtran}
\IEEEoverridecommandlockouts
\usepackage{cite}
\usepackage{amsmath,amssymb,amsfonts}
\usepackage{algorithmic}
\usepackage{graphicx}
\usepackage{textcomp}
\usepackage{bm,color, xcolor, mathtools, multirow, tabularx}
\usepackage{subcaption}
\ifdefined\ARXIV\else\usepackage{epstopdf}\fi
\usepackage{balance}
\newcommand{\Figref}[1]{{\bf Fig.\hspace{2pt}\ref{#1}}}
\newcommand{\Figsref}[1]{{\bf Figs.\hspace{2pt}\ref{#1}}}
\newcommand{\Figsrefs}[1]{{\bf \ref{#1}}}
\newcommand{\Tabref}[1]{{\bf Table\hspace{2pt}\ref{#1}}}

\definecolor{Iblue}{rgb}{0.0,0.0,0.7}
\definecolor{Ired}{rgb}{0.7,0.0,0.0}

\DeclareMathOperator*{\argmax}{arg\,max}

\def\BibTeX{{\rm B\kern-.05em{\sc i\kern-.025em b}\kern-.08em
    T\kern-.1667em\lower.7ex\hbox{E}\kern-.125emX}}

\ifdefined\ARXIV
  \newcommand{\ieeenotice}{\thanks{\ieeenoticetext}}
\else
  \newcommand{\ieeenotice}{}
\fi

\begin{document}
\title{Impacts of Single-objective Landscapes on Multi-objective Optimization\\
}
\author{\IEEEauthorblockA{Shoichiro Tanaka $^\dagger$, Keiki Takadama $^\dagger$, and Hiroyuki Sato $^\dagger$}
\IEEEauthorblockA{$^\dagger$
	\textit{Graduate School of Information and Engineering Sciences} \\
\textit{The University of Electro-Communications}\\
1-5-1 Chofugaoka, Chofu, Tokyo, 182-8585 JAPAN}
Email: \{stanaka, keiki@inf, h.sato\}@uec.ac.jp
\ieeenotice
}

\maketitle

\begin{abstract}
This work revealed a relationship between a multi-objective optimization problem and single-objective optimization problems that exist in the multi-objective problem. This work focused on combinatorial problems and investigated the relations between the local optima networks of the single-objective problems and the Pareto optima network of the multi-objective problem. Each of their networks has a graph structure. We divided the entire network into subgraphs. Each subgraph was called a component and characterized by overlapping relations between the single-objective local optima networks and the multi-objective Pareto optima network. Results on multi-objective landscape problems showed that most Pareto optimal solutions were reachable from the single-objective local optimal solutions. This tendency was emphasized by increasing the number of objectives and the objective correlation. The number of co-variables impacted the number of cross-link relations between the single-objective local optima networks and the multi-objective Pareto optima network. The results suggested that searching for single-objective problems is a clue to multi-objective optimization.
\end{abstract}

\begin{IEEEkeywords}
local optima network, Pareto graph, multi-objective optimization, combinatorial optimization
\end{IEEEkeywords}

\section{Introduction}
Evolutionary algorithms are suited for solving multi-objective optimization problems involving multiple single-objective functions \cite{coello2007}. The task is to find the Pareto optimal solutions, representing the optimal trade-off of objective function values in the objective space. The multi-objective optimization tries to simultaneously acquire multiple points in the variable space, which are the Pareto optimal solutions. The distribution of the Pareto optimal solutions in the variable space influences the multi-objective optimization performance. However, we have difficulty in visualizing and analyzing the Pareto optimal solutions in a high-dimensional variable space. Characteristics of multi-objective optimization problems are commonly discussed by using the Pareto front, which is a projection of the Pareto optimal solutions in the variable space to the objective space \cite{tuvsar2014}. This method is useful also for multi-criteria decision-making, selecting one solution from the Pareto optimal solutions. However, it is quite different from representing the distances between the Pareto optimal solutions in the variable space and the findability of the Pareto optimal solutions.

The local optima network (LON) \cite{ochoa2008} is a landscape representation of a target single-objective combinatorial optimization problem in the variable space. 
LON is a graph. Nodes are local optimal solutions, and edges are transitions between local optimal solutions by an optimization algorithm. A similar concept has been used also for multi-objective optimization problems so far \cite{paquete2009,liefooghe2018,liefooghe2019}. In the Pareto graph \cite{paquete2009}, each node is a Pareto optimal solution, and each edge represents that connected two Pareto optimal solutions are reachable to each other by the Pareto dominance based local search. In PLOS-Net \cite{liefooghe2018}, each node is a local Pareto optimal solution, and each edge represents that connected two local Pareto optimal solutions are reachable to each other by the Pareto dominance based local search. In both the Pareto graph and PLOS-Net, a solution set connected by edges becomes a subgraph also called a component. These previous works surveyed components of the Pareto optimal solutions \cite{paquete2009} and the local Pareto optimal solutions \cite{liefooghe2018} from the viewpoints of the component size and the number of components in the graph. On the other hand, this work focuses on the Pareto optimal solution network (PON), mentioned as the Pareto graph \cite{paquete2009} and investigates the relations between the multi-objective PON components and the single-objective LONs involved in the target multi-objective optimization problem. 

This work aims to classify the multi-objective PON components based on the relations with the single-objective LONs and investigate the relations between the multi-objective PON and the single-objective LONs simultaneously exist in the target multi-objective combinatorial optimization problem. We target $\rho MNK$ landscape problems \cite{verel2011_2} with different objective correlations $\rho$, numbers of objectives $M$, and numbers of co-variables $K$. We discuss the distribution of the Pareto optimal solutions in the variable space and their findability by considering relations with the underlying single-objective problems in the target multi-objective problem.

\section{Definitions}
\subsection{$\rho MNK$ Landscape}
$\rho MNK$ landscape \cite{verel2011_2} is a multi-objective combinatorial problem framework, which can vary the objective correlation $\rho$, the number of objectives $M$, the number of variables $N$, and the number of co-variables $K$. The single-objective landscape \cite{kauffman1987} was extended to the $MNK$ landscape \cite{aguirre2004,allmendinger2022} in order to involve $M$ objectives. $MNK$ landscape was further extended to the $\rho MNK$ landscape in order to vary the objective correlations among $M$ objectives. The variable space of $\rho MNK$ landscape problems is given by $\mathcal{X}=\{0,1\}^N $. For variable vector $\bm x=(x_1,x_2,\dotsc,x_N)\in\mathcal{X}$, $\rho MNK$ landscape problems is given by
\begin{multline}
	{\rm Maximize} \ f_i(\bm{x})\ (i=1,\ldots,M),\\
	f_i(\bm x) = \frac{1}{N}\sum^{N}_{j=1}g_{ij}(x_j, x_{j1}, \ldots, x_{jK}), \label{f_sub}
\end{multline}
where, $g_{ij}$ $(j=1,\ldots,N)$ are sub-functions for $i$-th objective function $f_i$. Variable $x_j$ and $K$ co-variables $\{x_{j1}, \ldots, x_{jK}\}$ determine the value of sub-function $g_{ij}$. $K$ co-variables are randomly chosen from all $N$ variables. The optimization difficulty generally increases as the number of co-variables $K$ increases. Each sub-function $g_{ij}$ uses a multivariate uniform distribution with a $M\times M$ positive definite symmetric covariance matrix $O$, i.e. $o_{i, i}=1$ and $o_{i, j}=\rho$ for all $i, j \in \{1, \ldots, M\}$. Correlations between objectives consequently becomes $\rho > \frac{-1}{M-1}$.

\subsection{Essentials in Single-objective Optimization}
The $\epsilon$-neighbor solution set of solution $\bm{x}$ is represented as $\mathcal{N}_{\epsilon}(\bm{x})= \{ \bm{x}' \in \mathcal{X}\backslash \{\bm{x}\} \mid d(\bm{x}, \bm{x}') \leq \epsilon \}$, where $d(\bm{x}, \bm{x'})$ is the Hamming distance between solutions $\bm{x}$ and $\bm{x'}$ in the variable space $\mathcal{X}$. For single-objective function $f$, the local optimal solution set $L$ is given by
\begin{equation}
	L=\{\bm{x} \in \mathcal{X} \mid \nexists \bm{x}' \in \mathcal{N}_{\epsilon=1}(\bm{x}):f(\bm{x}) < f(\bm{x}') \}.
\end{equation}
The global optimal solution is given by $\argmax_{\bm x \in \mathcal{X}}f(\bm x)$.

\subsection{Essentials in Multi-objective Optimization}
For two solutions $\bm{x},\bm{x'}\in \mathcal{X}$ and $M$ dimensional objective function vector $\bm f=(f_1,\ldots,f_M)$, $\bm{x}$ dominates $\bm{x'}$ ($\bm{x} \succ \bm{x'}$) if and only if
\begin{align}
	\begin{array}{ll}
		\forall m \in \{1, \ldots, M\}: f_m(\bm{x}) \geq f_m(\bm{x'}) & \land \\
		\exists m \in \{1, \ldots, M\}: f_m(\bm{x}) > f_m(\bm{x'}). & ~
	\end{array}
\end{align}
Solution $\bm{x}\in\mathcal{X}$ is said to be local Pareto optimal when $\bm{x}$ is non-dominated in its $\epsilon=1$ neighbor solution set $\mathcal{N}_{\epsilon=1}(\bm{x})$. The local Pareto optimal solution set is then given by $LP=\{\bm{x} \in \mathcal{X} \mid \nexists\bm{x'} \in \mathcal{N}_{\epsilon=1}(\bm{x}):\bm{x'} \succ \bm{x}\}$.
Also, the Pareto optimal solution set is given by $P=\{\bm{x} \in \mathcal{X} \mid \nexists\bm{x'} \in \mathcal{X}:\bm{x'} \succ \bm{x}\}$ as the non-dominated solution set in the variable space $\mathcal{X}$.

\section{Local optima networks}
\subsection{Method}
The local optima network (LON) is a landscape representation methodology of single-objective combinatorial optimization problems \cite{ochoa2008}. LON represents search transitions of an optimization algorithm on the target single-objective problem by a directed graph $G_{LON}=(L, E)$. Node set $L$ is the local optimal solution set $L$. Edge set $E$ is the transition set between local optimal solutions. There are several options in the edge representation. This work uses the monotonic perturbation edge \cite{ochoa2017}. For two local optimal solutions $\bm{l}, \bm{l'} \in L$, we generate an edge from $\bm{l}$ to $\bm{l}'$ when $\bm{l'}$ is reached by the local search from a solution perturbed from $\bm{l}$. The perturbation is set to less than three bits in this work. The weight of the monotonic perturbation edge between $\bm{l}$ and $\bm{l'}$ is the transition probability from $\bm{l}$ to $\bm{l}'$ \cite{ochoa2017}.
\subsection{Example}
\begin{figure}[t]
	\begin{center}
		\includegraphics[width=0.38\linewidth]{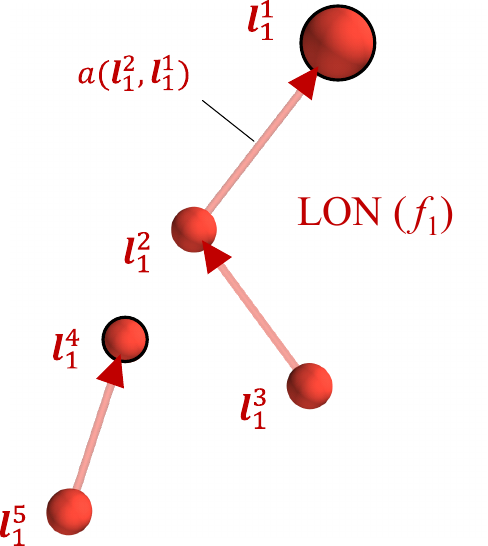}
	\end{center}
	\caption{The single-objective LON}
	\label{fig:LON1}
\end{figure}

\Figref{fig:LON1} shows an example LON of single-objective function $f_1$. The node set $L$ is the local optimal solution set $L=\{\bm l^1_1,\bm l^2_1,\bm l^3_1,\bm l^4_1,\bm l^5_1\}$. The arc set $A$ is the transition set $A=\{a(\bm l^3_1,\bm l^2_1),a(\bm l^2_1,\bm l^1_1),a(\bm l^5_1,\bm l^4_1)\}$ between local optimal solutions. For example, arc $a(\bm l^2_1,\bm l^1_1)$ represents that the optimization algorithm, the perturbation local search in this work, finds local optimal solution $\bm l^1_1$ from $\bm l^2_1$. Local optimal solutions $\bm l^1_1$ and $\bm l^4_1$ not transiting to any other local optimal solutions are shown in black. Each of them is called the {\it sink}. Local optimal solution set converged to a sink is called the {\it funnel}. The funnel of sink $\bm l^1_1$ is $\{\bm l^1_1,\bm l^2_1,\bm l^3_1\}$. The funnel of sink $\bm l^4_1$ is $\{\bm l^4_1,\bm l^5_1\}$. \Figref{fig:LON1} shows a case that the single-objective function $f_1$ has two funnels.
The funnel in \Figref{fig:LON1} is tiny for simplicity. Usually, a funnel contains many more local optimal solutions.

\begin{figure}[t]
	\begin{center}
		\includegraphics[width=0.8\linewidth]{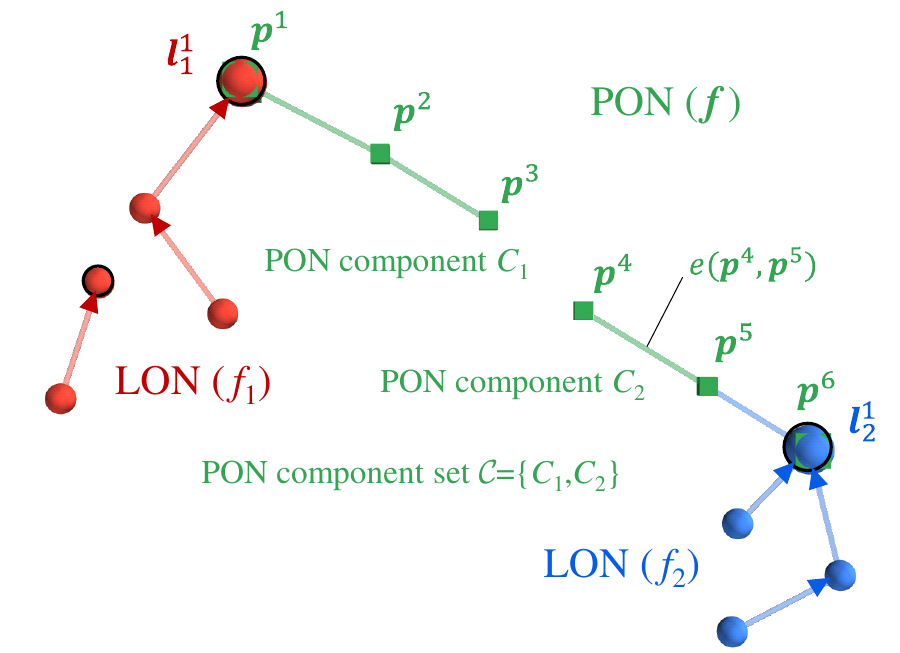}
	\end{center}
	\caption{The single objective LONs and the multi-objective PON}
	\label{fig:PON1}
\end{figure}

\begin{figure*}[t]
	\begin{center}
		\includegraphics[width=1\linewidth]{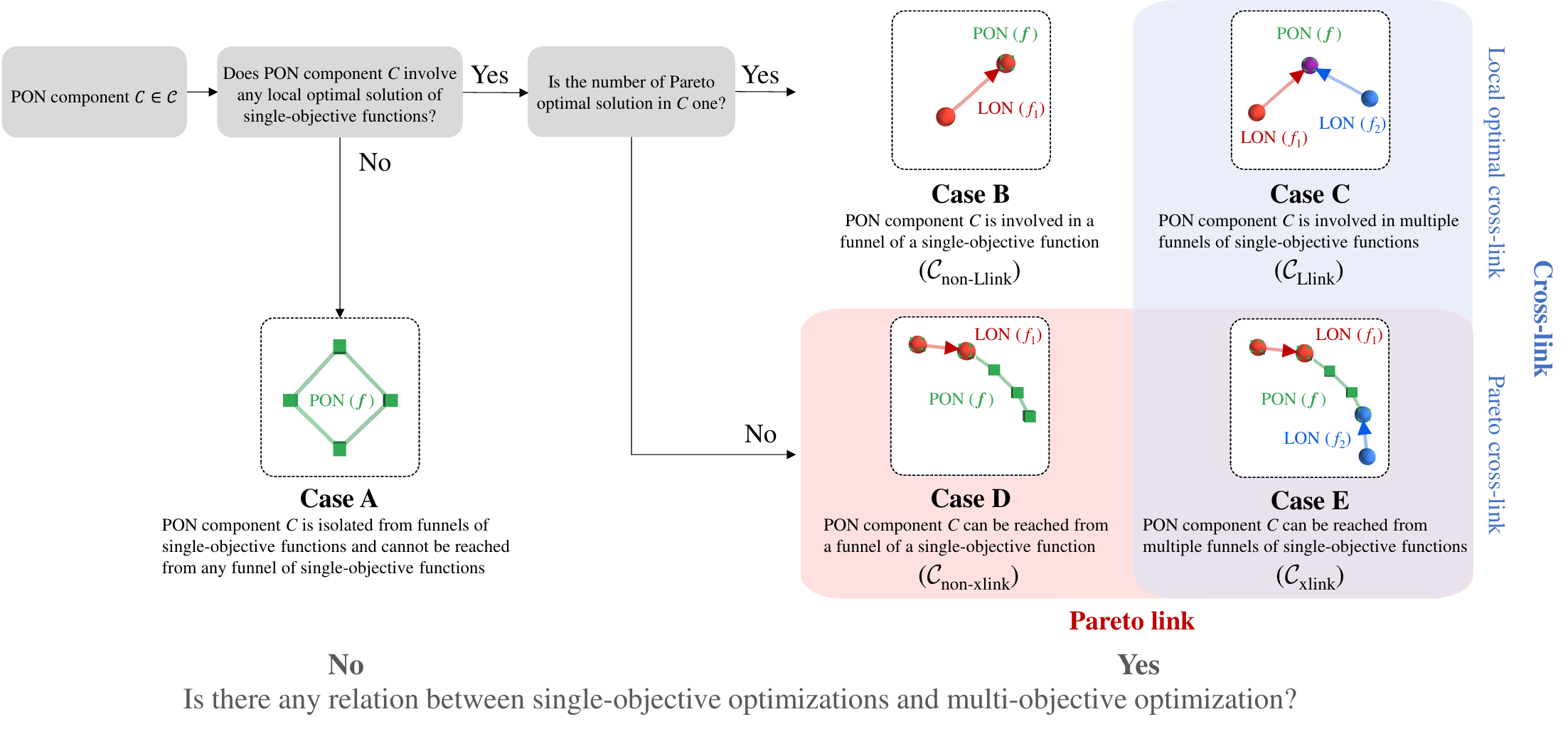}
	\end{center}
	\caption{Classification of a multi-objective PON component $C \in \mathcal{C}$ based on relations with the single-objective LONs}
	\label{fig:LONPON_CASES}
\end{figure*}

\section{Pareto Optima Network}
\subsection{Method}
The Pareto optima network (PON) also mentioned as the Pareto graph \cite{paquete2009} represents the relations between the Pareto optimal solutions in an undirected graph $G_{PON}(P, E)$. Node set $P$ is the Pareto optimal solutions $P$. Edge set $E$ is the transition set between Pareto optimal solutions. For two Pareto optimal solutions $\bm{p}, \bm{p}' \in P$, we generate an edge between them when they are reachable to each other by the one-bit-flip mutation.

\subsection{Example}
\Figref{fig:PON1} shows a PON of two-dimensional objective function vector  $\bm f=(f_1,f_2)$ in green. Node set is the Pareto optimal solution set $P=\{\bm p^1,\bm p^2,\bm p^3,\bm p^4,\bm p^5,\bm p^6\}$. Edge set is the transitions $E=\{e(\bm p^1,\bm p^2),e(\bm p^2,\bm p^3),e(\bm p^4,\bm p^5),e(\bm p^5,\bm p^6)\}$ between the Pareto optimal solutions. For example, $e(\bm p^4,\bm p^5)$ indicates that two Pareto optimal solutions $\bm p^4$ and $\bm p^5$ are mutually reachable by the one-bit-flip mutation.

\section{Proposal: PON Component Classification Based on Relations with Single-objective LONs}
This work classifies the components of the multi-objective PON and investigate the relations between each PON component and the single-objective LONs. 

\subsection{PON component}
PON can be divided into subgraphs, and each of them are called a PON component. PON in \Figref{fig:PON1} has two components $C_1$ and $C_2$. The first component is $C_1=\{\bm p^1,\bm p^2,\bm p^3\}$. The second component is $C_2=\{\bm p^4,\bm p^5,\bm p^6\}$. Any solution in $C_1$ cannot reach to any solution in $C_2$ by the one-bit-flip mutation. Also, any solution in $C_2$ cannot reach to any solution in $C_1$ by the one-bit-flip mutation. There is no edge between $C_1$ and $C_2$, and they are recognized as different PON components. The PON component set $\mathcal{C}$ is the set of all PON components. In \Figref{fig:PON1}, the PON component set is $\mathcal{C}=\{C_1, C_2\}$.

\subsection{Relation of PON Component and Single-objective LONs}
\Figref{fig:PON1} also involves the single-objective LON of the objective function $f_1$ in red and the single-objective LON of the objective function $f_2$ in blue. The same solution in different networks is overlapped in \Figref{fig:PON1}. For example, the black global optimal solution $\bm l^1_1$ of the objective function $f_1$ is overlapped with the green Pareto optimal solution $\bm p^1$, i.e. $\bm l^1_1=\bm p^1$. The black global optimal solution $\bm l^1_2$ is also overlapped with the green Pareto optimal solution $\bm p^6$, i.e. $\bm l^1_2=\bm p^6$. These overlaps between the single-objective LONs and the multi-objective PON indicate relations between the single-objective LONs and the multi-objective PON.

\subsection{PON Component Classification Criteria}
We classify each PON component $C\in \mathcal{C}$ into cases A--E in the procedure shown in \Figref{fig:LONPON_CASES}. We describe each of the six cases A--E below.

Case A is that the target PON component is isolated from single-objective LONs. In this case, the PON component does not have any relation with single-objective LONs. Cases B and C are that the target PON component has only one Pareto optimal solution. Case B is that a single-objective function's funnel involves the sole Pareto optimal solution of the target PON component. Case C is that multiple objective functions' funnels involve the sole Pareto optimal solution of the target PON component. Cases D and E are that the target PON component has multiple Pareto optimal solutions. Case D is that a single-objective function's funnel involves one or more Pareto optimal solutions of the target PON component. Case E is that multiple objective functions' funnels involve one or more Pareto optimal solutions of the target PON component.

In cases D and E, the PON component is reachable from at least one objective function's funnel. This kind of link connecting a PON component and at least one objective function's funnel is called the {\it Pareto-link} in this work and shown in the red area in \Figref{fig:LONPON_CASES}. In cases C and E, multiple single-objective functions' funnels are mutually reachable. This kind of link connecting multiple single-objective functions' funnels is called the {\it cross-link} in this work and shown in the blue area in \Figref{fig:LONPON_CASES}. Especially, case C is called the {\it local optima cross-link} since multiple single-objective functions' funnels cross-link via local optimal solutions even without the PON component. On the other hand, case E is called {\it Pareto optima cross-link} since multiple single-objective functions' funnels cross-link via the PON component.

In order to analyze the characteristics of PON components, classified components to cases B--E are respectively inputted to component sets named $\mathcal{C}_\text{non-Llink}$, $\mathcal{C}_\text{Llink}$, $\mathcal{C}_\text{non-xlink}$, $\mathcal{C}_\text{xlink}$ shown in \Figref{fig:LONPON_CASES}.

\begin{table}[t]
	\caption{Quantitative metrics of the multi-objective PON and the single-objective LONs}
	\label{tab:indicator}
	{\renewcommand\arraystretch{1}
		\begin{tabularx}{\linewidth}{l|X }
			\multicolumn{1}{c|}{Name}                 & \multicolumn{1}{c}{Description} \\
			\hline \hline
			$n_\text{Pareto}$                  & The number of Pareto optimal solutions $|P|$ in the target problem instance, the number of nodes in the multi-objective PON of the target problem instance.\\ 
			\hline
			$n_\text{comp}$                    & The number of PON components $|\mathcal{C}|$ in the multi-objective PON of the target problem instance.\\
			\hline
			$n_\text{funnel}$                  & The total number of funnels (sinks) in all $M$ single-objective functions of the target problem instance.\\
			\hline
			$n_\text{xlink}$                   &
			The number of PON components classified into case C, $|\mathcal{C}_\text{xlink}|$. \\
			\hline
			$r_\text{link}$                    & 
			The ratio of the Pareto optimal solutions in PON components classified into cases D and E (Pareto link) to all Pareto optimal solutions $P$, $\sum_{C\in \mathcal{C}_\text{non-xlink}\cup \mathcal{C}_\text{xlink}}|C|/|P|$.\\
			\hline
			$r_\text{xlink}$                   & 
			The ratio of Pareto optimal solutions in PON components classified into case E (Pareto cross-link) to all Pareto optimal solutions $P$, $\sum_{C\in \mathcal{C}_\text{xlink}}|C|/|P|$.\\
			\hline
			$r_\text{non-xlink}$     & 
			The ratio of Pareto optimal solutions in PON components classified into case D to all Pareto optimal solutions $P$, $\sum_{C\in \mathcal{C}_\text{non-xlink}}|C|/|P|$.\\
			\hline
			$r_\text{Llink}$                   & 
			The ratio of Pareto optimal solutions in components classified into case C (Local optima cross-link) to all Pareto optimal solutions $P$, $\sum_{C\in \mathcal{C}_\text{Llink}}|C|/|P|$.
		\end{tabularx}
	}
\end{table}

\section{Experimental Setup}
We qualitatively and quantitatively investigated the impacts of the single-objective LONs based on the single-objective functions $f_i$ ($i=1,\ldots,M$) on the multi-objective PON based on the multi-objective function vector $\bm f=(f_1,\ldots,f_M)$.

\subsection{Problem}
We used $\rho MNK$ landscape problems with the objective correlations $\rho\in\{-0.8, -0.6, -0.4, -0.2, 0.0, 0.2, 0.4, 0.6,$ $ 0.8\}$, the number of objectives $M\in\{2,3,4\}$, the number of variables $N=18$, and the number of co-variables $K\in\{1,2,3,4\}$. 
We generated 50 different and independent problem instances for each parameter combination (3,300 instances in total) to suppress the influence of the random creation. To fully cover the problem characteristics of each problem instance, we conducted the exhaustive brute force search to obtain the local optimal solutions $L$ of each single objective function $f\in\{f_1,\ldots,f_M\}$ and the Pareto optimal solutions $P$ and generated the single-objective LONs and the multi-objective PON.

\subsection{Metric}
\Tabref{tab:indicator} shows metrics to quantitatively assess the single-objective LONs and the multi-objective PON.

\begin{figure*}[t]
	\begin{minipage}{0.245\textwidth}
		\begin{center}
			\includegraphics[width=0.95\textwidth,clip]{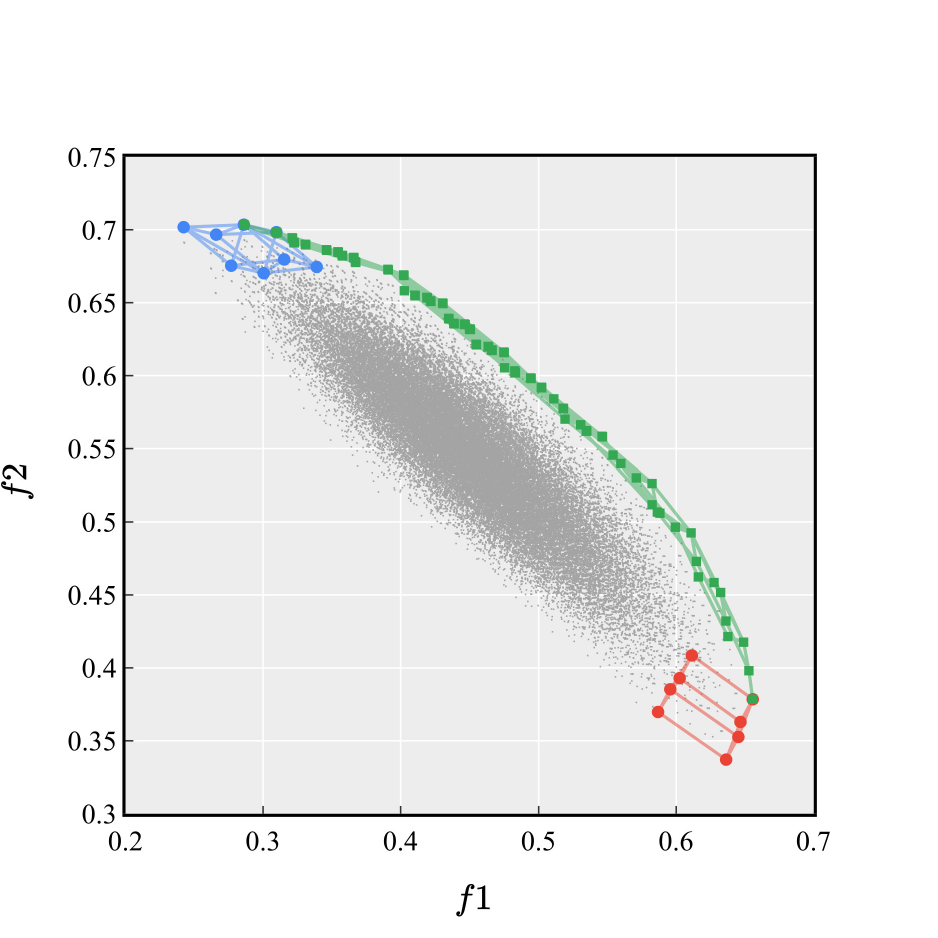}
			\\\vspace{-0.2cm}
			\subcaption{$K=1$}
		\end{center}
	\end{minipage}
	\begin{minipage}{0.245\textwidth}
		\begin{center}
			\includegraphics[width=0.95\textwidth,clip]{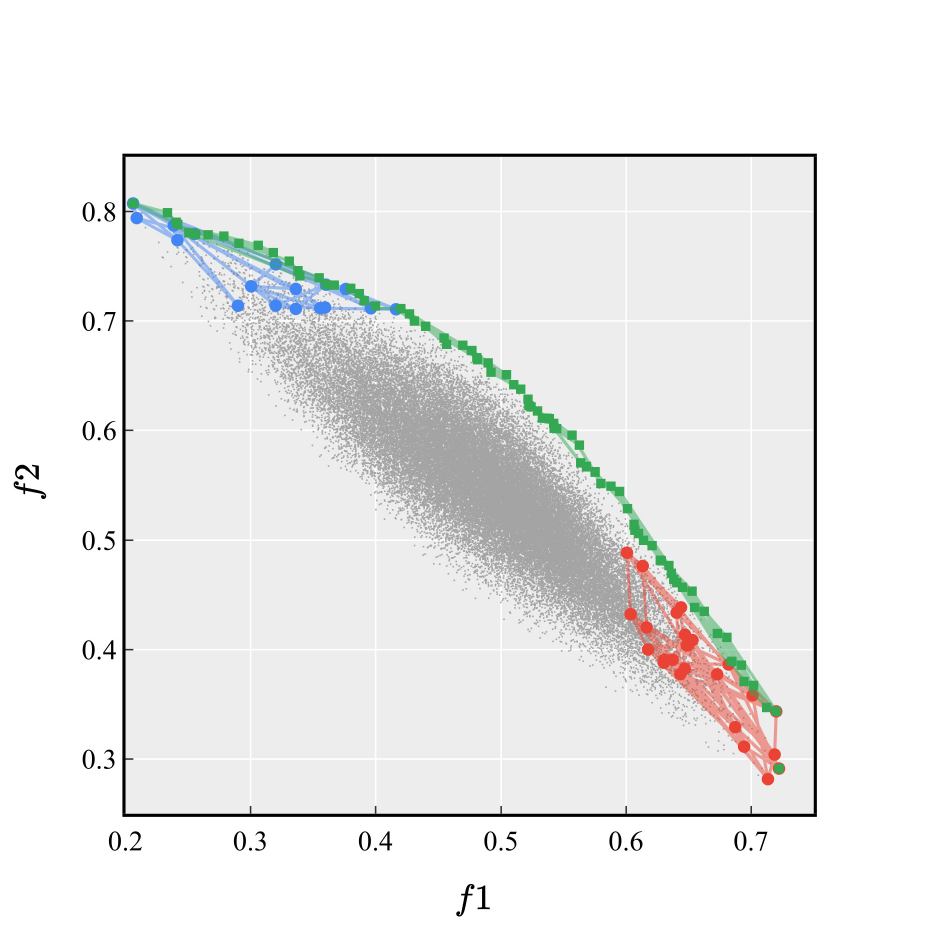}
			\\\vspace{-0.2cm}
			\subcaption{$K=2$}
		\end{center}
	\end{minipage}
	\begin{minipage}{0.245\textwidth}
		\begin{center}
			\includegraphics[width=0.95\textwidth,clip]{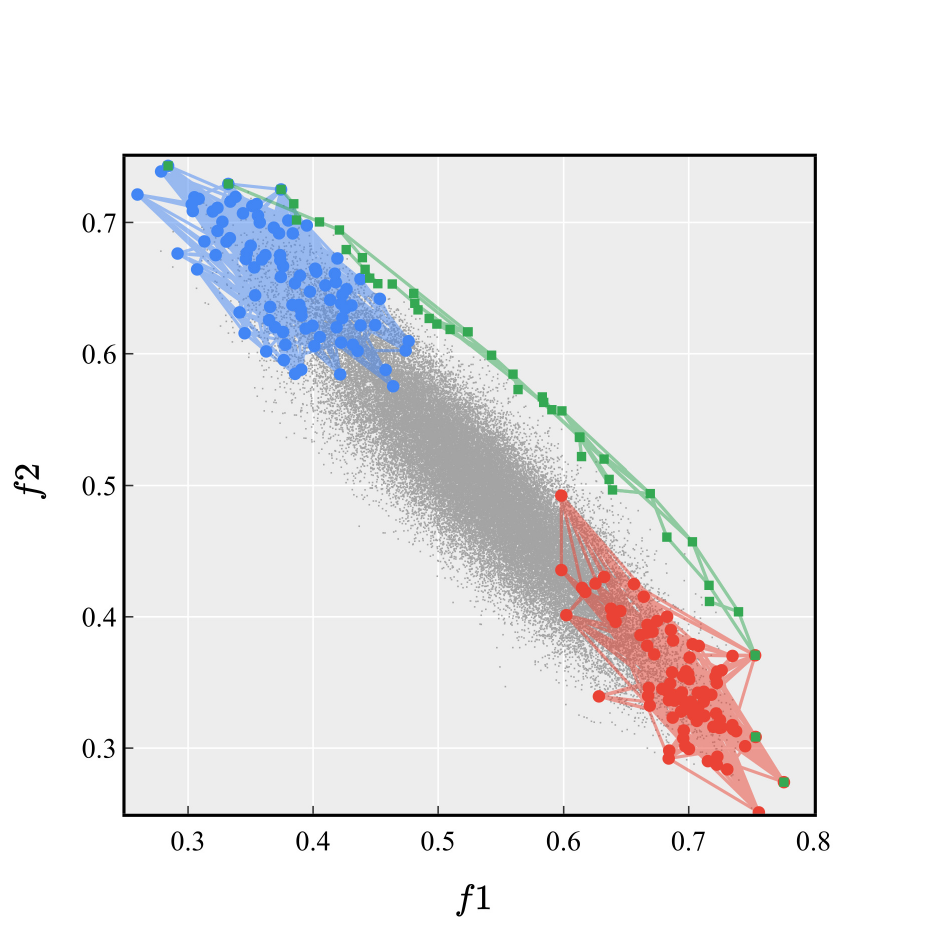}
			\\\vspace{-0.2cm}
			\subcaption{$K=3$}
		\end{center}
	\end{minipage}
	\begin{minipage}{0.245\textwidth}
		\begin{center}
			\includegraphics[width=0.95\textwidth,clip]{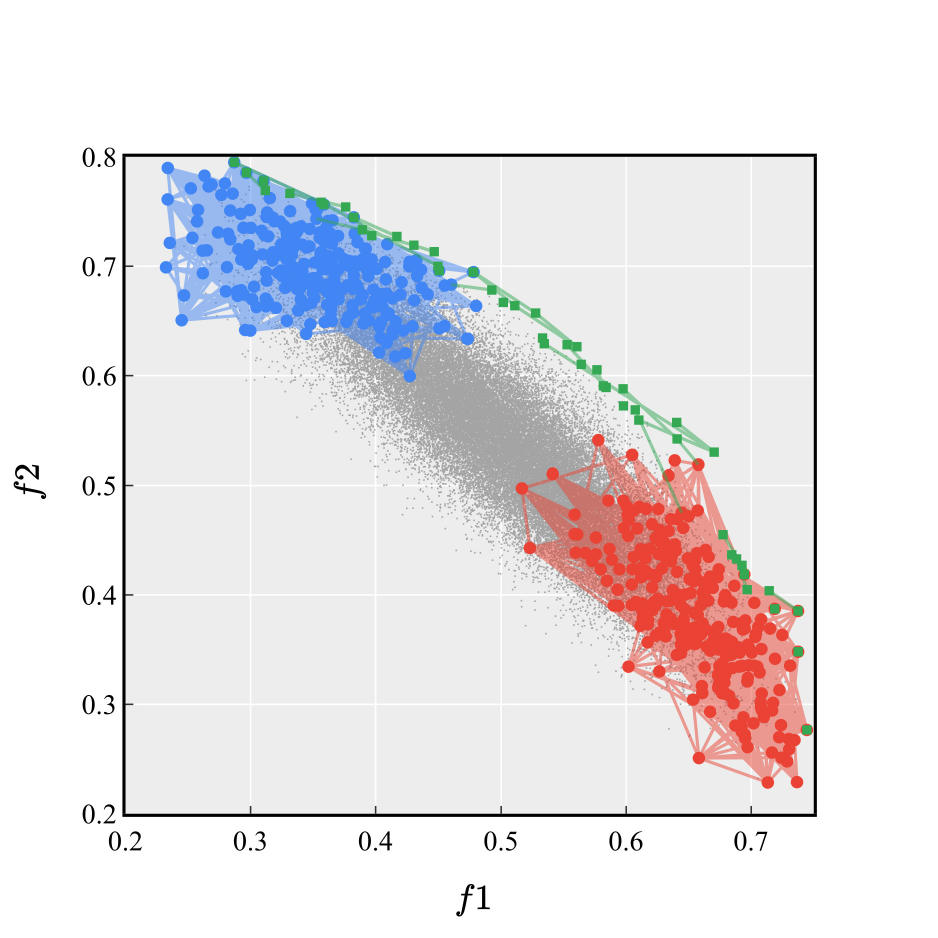}
			\\\vspace{-0.2cm}
			\subcaption{$K=4$}
		\end{center}
	\end{minipage}
	\vspace{-0.1cm}
	\caption{$M=2$ objectives, correlation $\rho=-0.8$, and different number of co-variables $K$}
	\label{fig:M2R-0.8}
	
	\vspace{-0.1cm}
	\begin{minipage}{0.245\textwidth}
		\begin{center}
			\includegraphics[width=0.95\textwidth,clip]{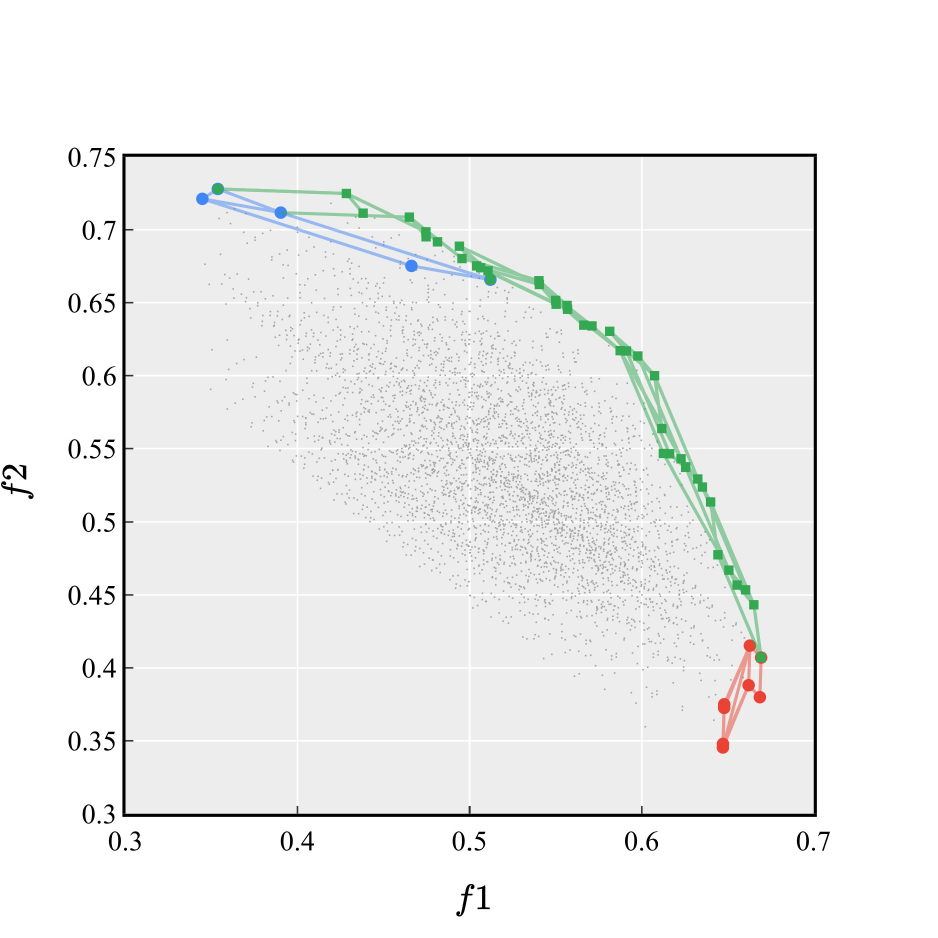}
			\\\vspace{-0.2cm}
			\subcaption{$K=1$}
		\end{center}
	\end{minipage}
	\begin{minipage}{0.245\textwidth}
		\begin{center}
			\includegraphics[width=0.95\textwidth,clip]{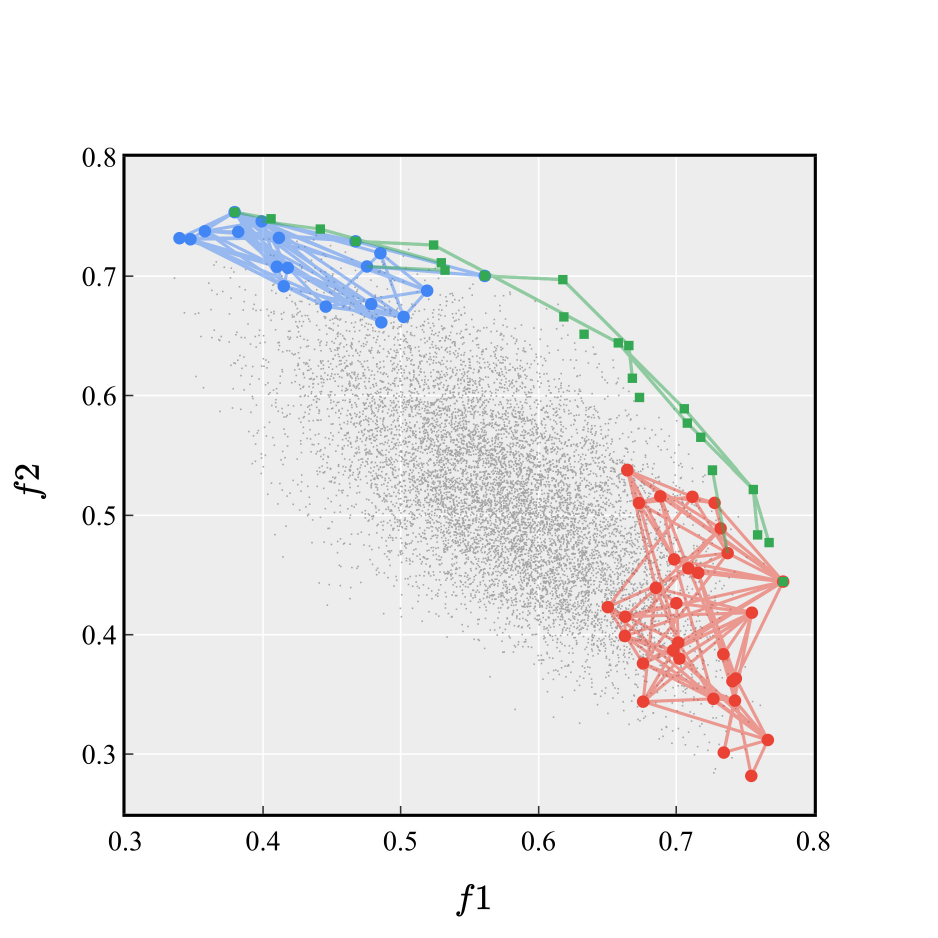}
			\\\vspace{-0.2cm}
			\subcaption{$K=2$}
		\end{center}
	\end{minipage}
	\begin{minipage}{0.245\textwidth}
		\begin{center}
			\includegraphics[width=0.95\textwidth,clip]{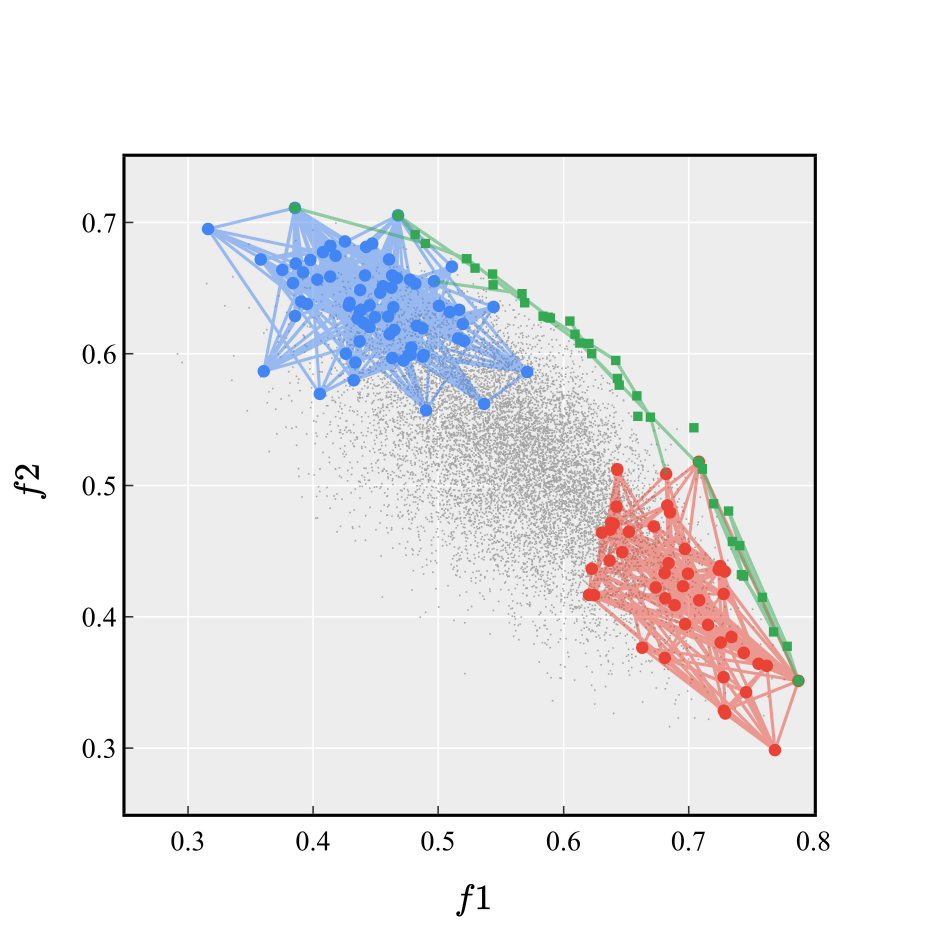}
			\\\vspace{-0.2cm}
			\subcaption{$K=3$}
		\end{center}
	\end{minipage}
	\begin{minipage}{0.245\textwidth}
		\begin{center}
			\includegraphics[width=0.95\textwidth,clip]{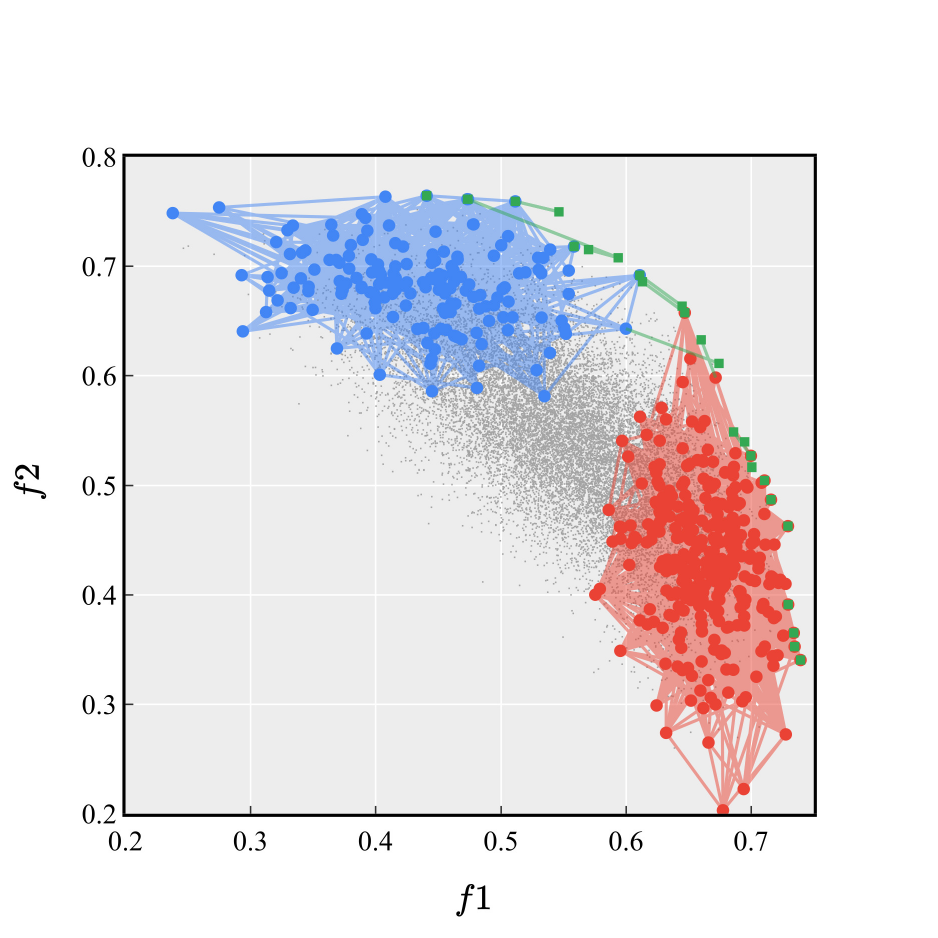}
			\\\vspace{-0.2cm}
			\subcaption{$K=4$}
		\end{center}
	\end{minipage}
\vspace{-0.1cm}
	\caption{$M=2$ objectives, correlation $\rho=-0.4$, and different number of co-variables $K$}
	\label{fig:M2R-0.4}
	
	\vspace{-0.1cm}
	\begin{minipage}{0.245\textwidth}
		\begin{center}
			\includegraphics[width=0.95\textwidth,clip]{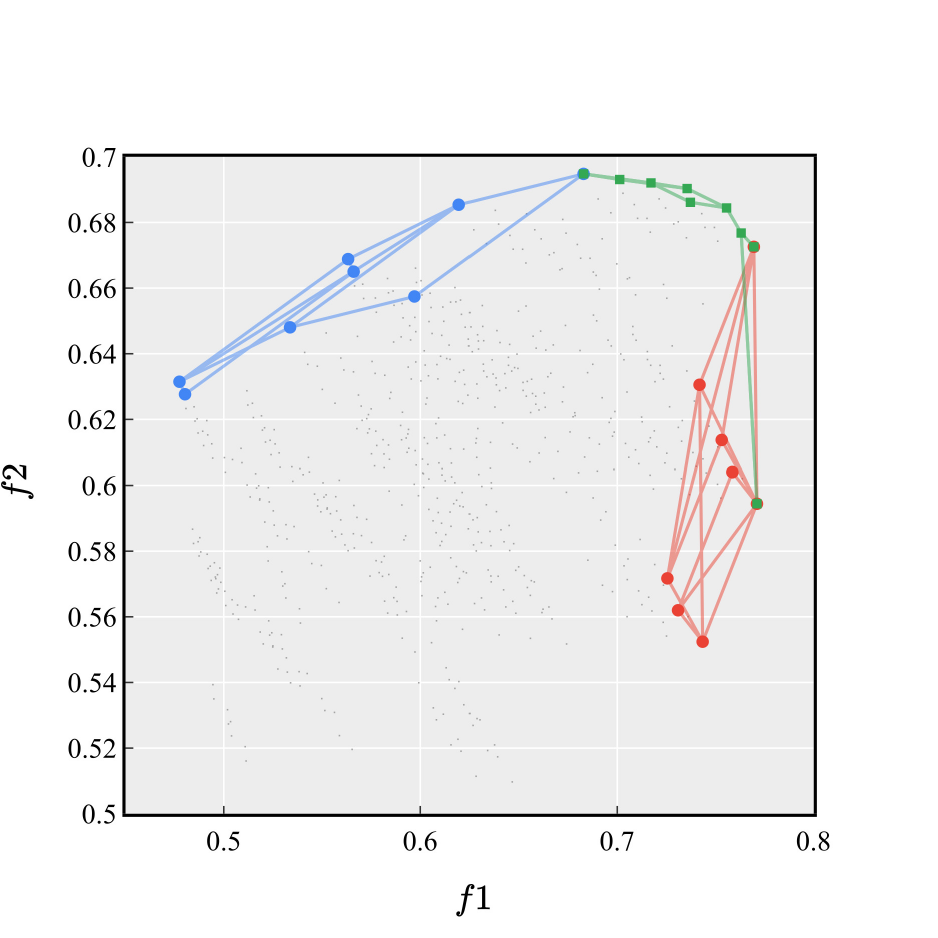}
			\\\vspace{-0.2cm}
			\subcaption{$K=1$}
		\end{center}
	\end{minipage}
	\begin{minipage}{0.245\textwidth}
		\begin{center}
			\includegraphics[width=0.95\textwidth,clip]{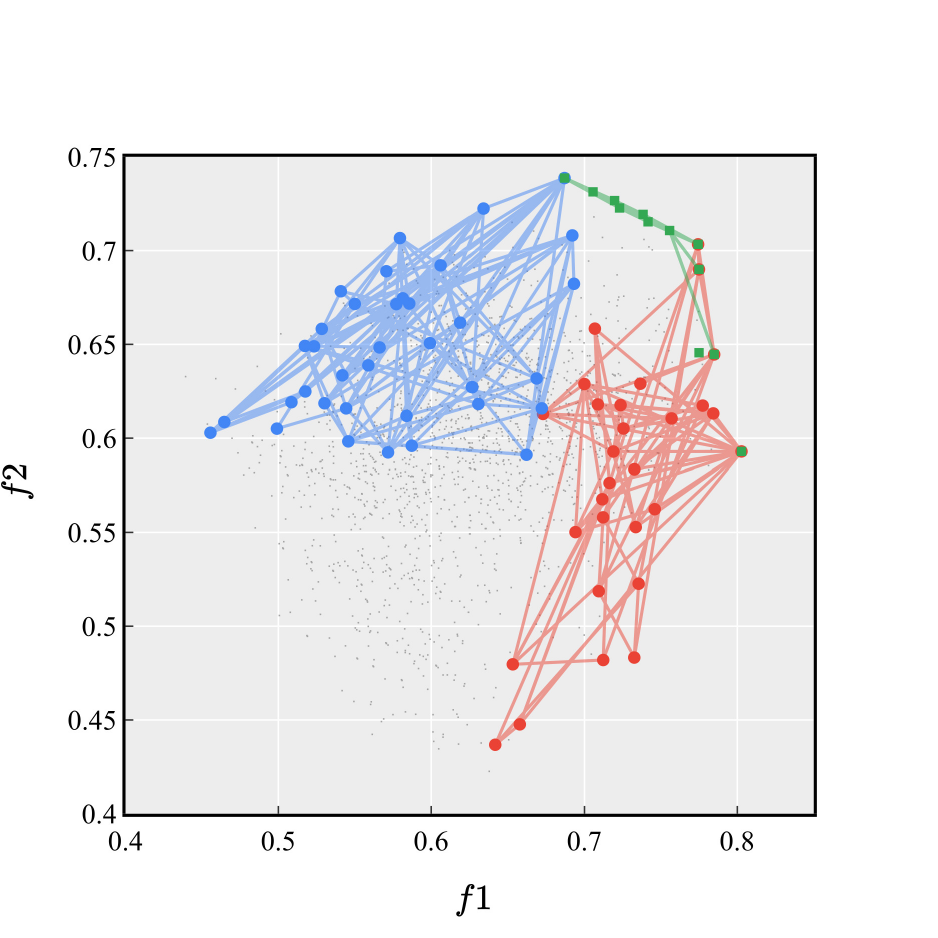}
			\\\vspace{-0.2cm}
			\subcaption{$K=2$}
		\end{center}
	\end{minipage}
	\begin{minipage}{0.245\textwidth}
		\begin{center}
			\includegraphics[width=0.95\textwidth,clip]{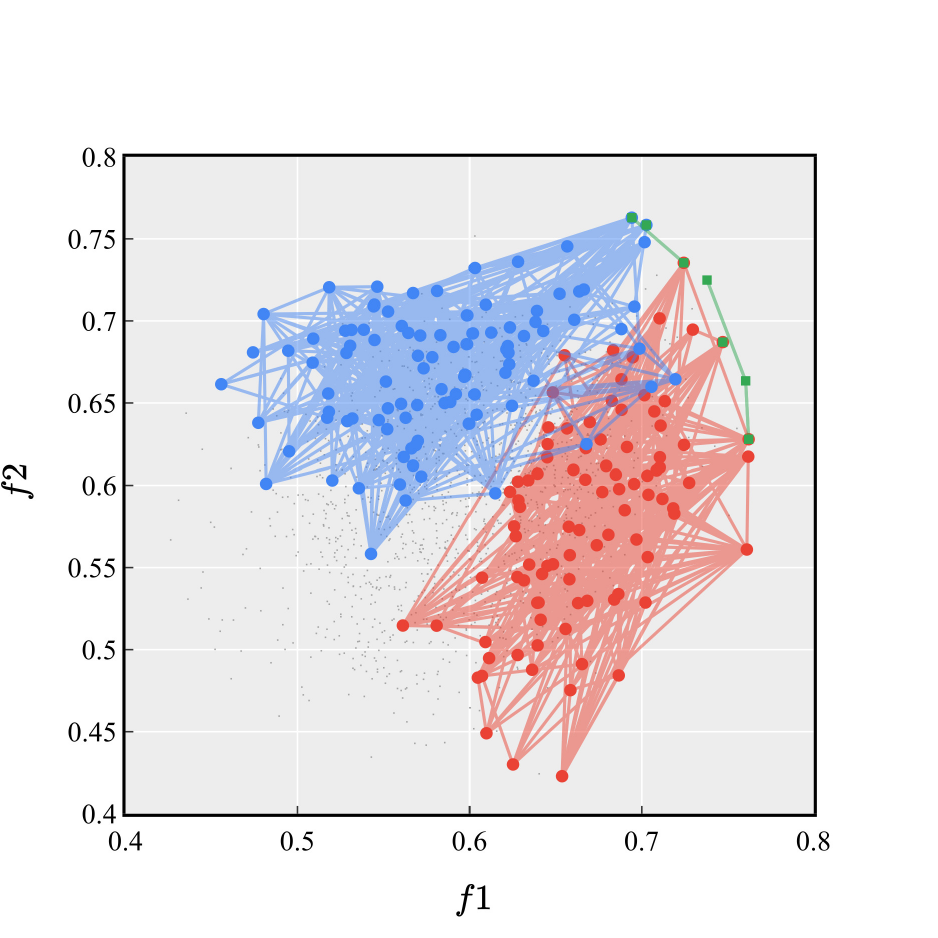}
			\\\vspace{-0.2cm}
			\subcaption{$K=3$}
		\end{center}
	\end{minipage}
	\begin{minipage}{0.245\textwidth}
		\begin{center}
			\includegraphics[width=0.95\textwidth,clip]{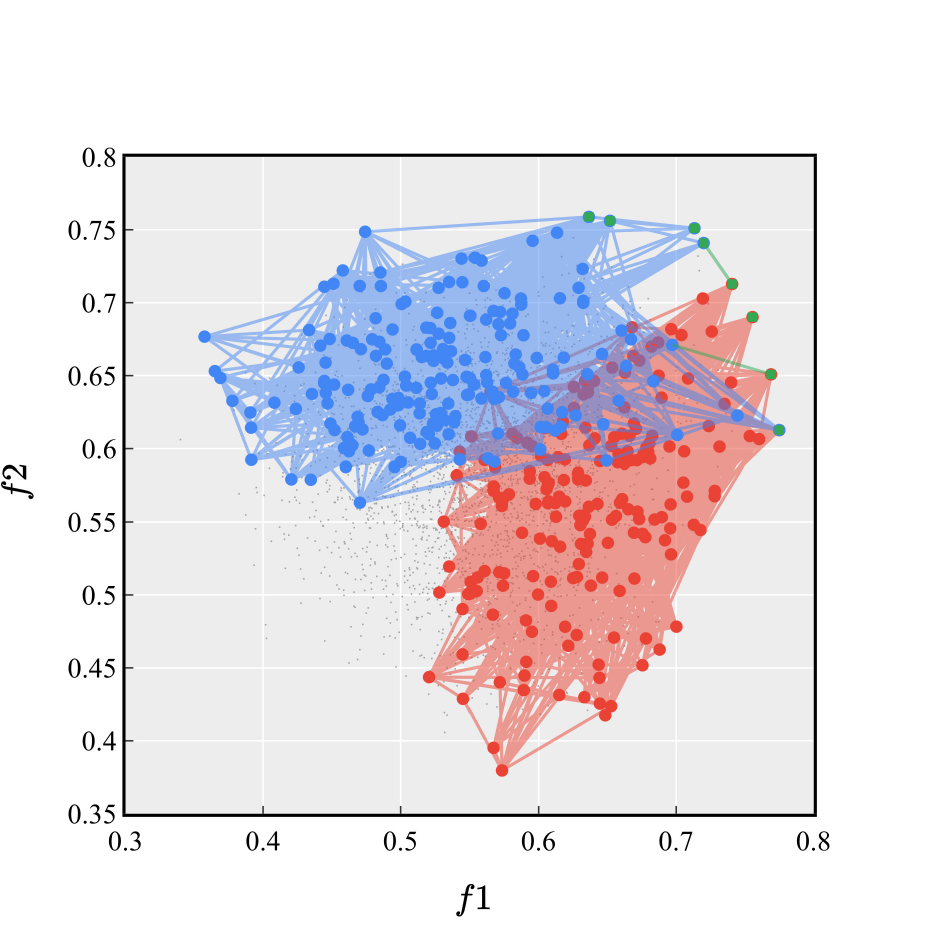}
			\\\vspace{-0.2cm}
			\subcaption{$K=3$}
		\end{center}
	\end{minipage}
\vspace{-0.1cm}
	\caption{$M=2$ objectives, correlation $\rho=0.4$, and different number of co-variables $K$}
	\label{fig:M2R0.4}
	
	\vspace{-0.1cm}
	\begin{minipage}{0.245\textwidth}
		\begin{center}
			\includegraphics[width=0.95\textwidth,clip]{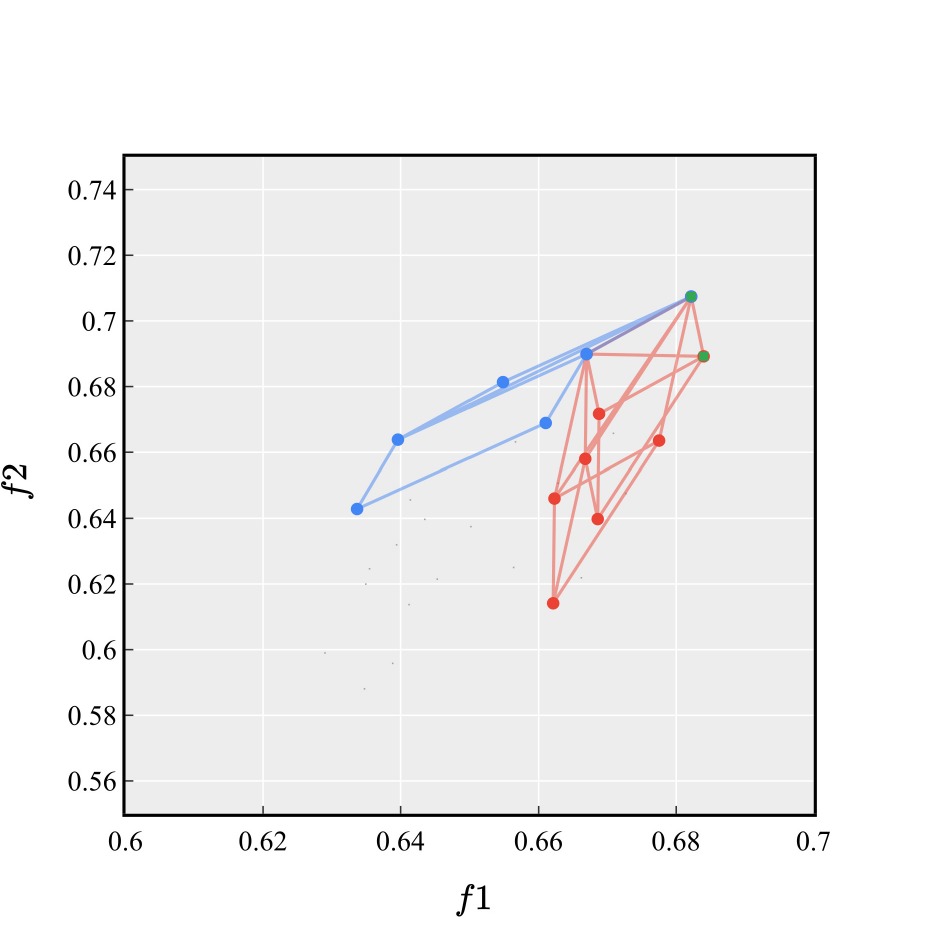}
			\\\vspace{-0.2cm}
			\subcaption{$K=1$}
		\end{center}
	\end{minipage}
	\begin{minipage}{0.245\textwidth}
		\begin{center}
			\includegraphics[width=0.95\textwidth,clip]{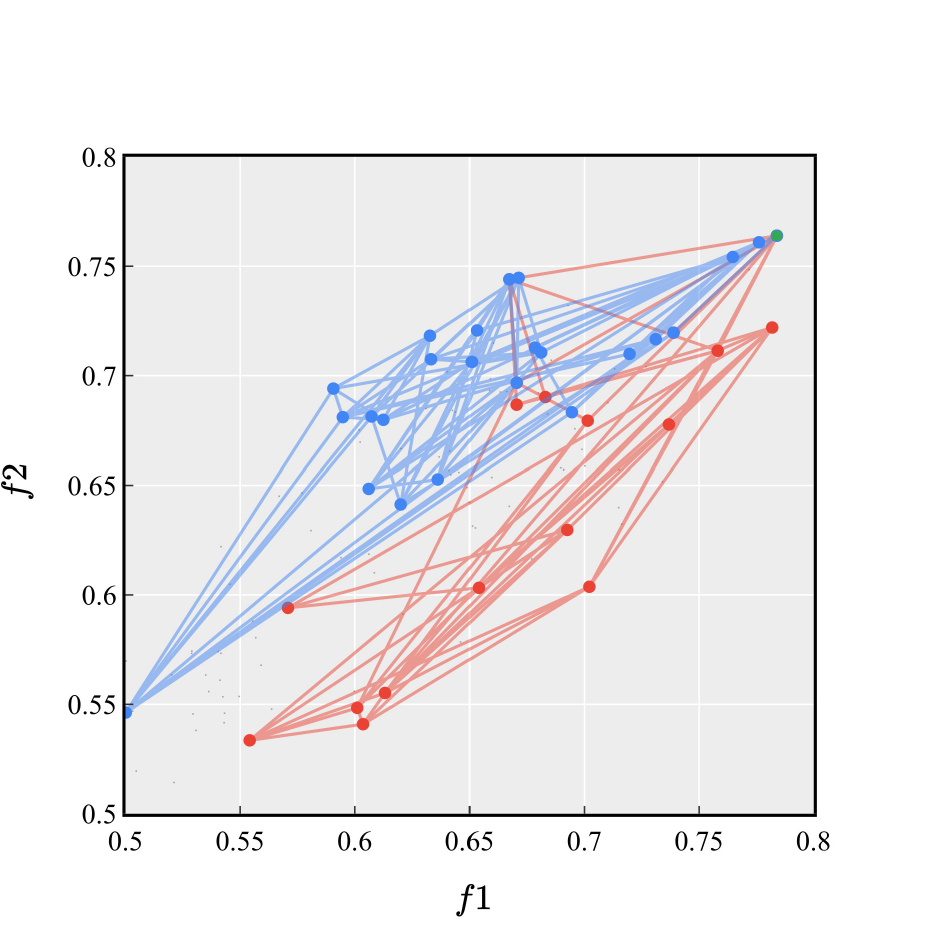}
			\\\vspace{-0.2cm}
			\subcaption{$K=2$}
		\end{center}
	\end{minipage}
	\begin{minipage}{0.245\textwidth}
		\begin{center}
			\includegraphics[width=0.95\textwidth,clip]{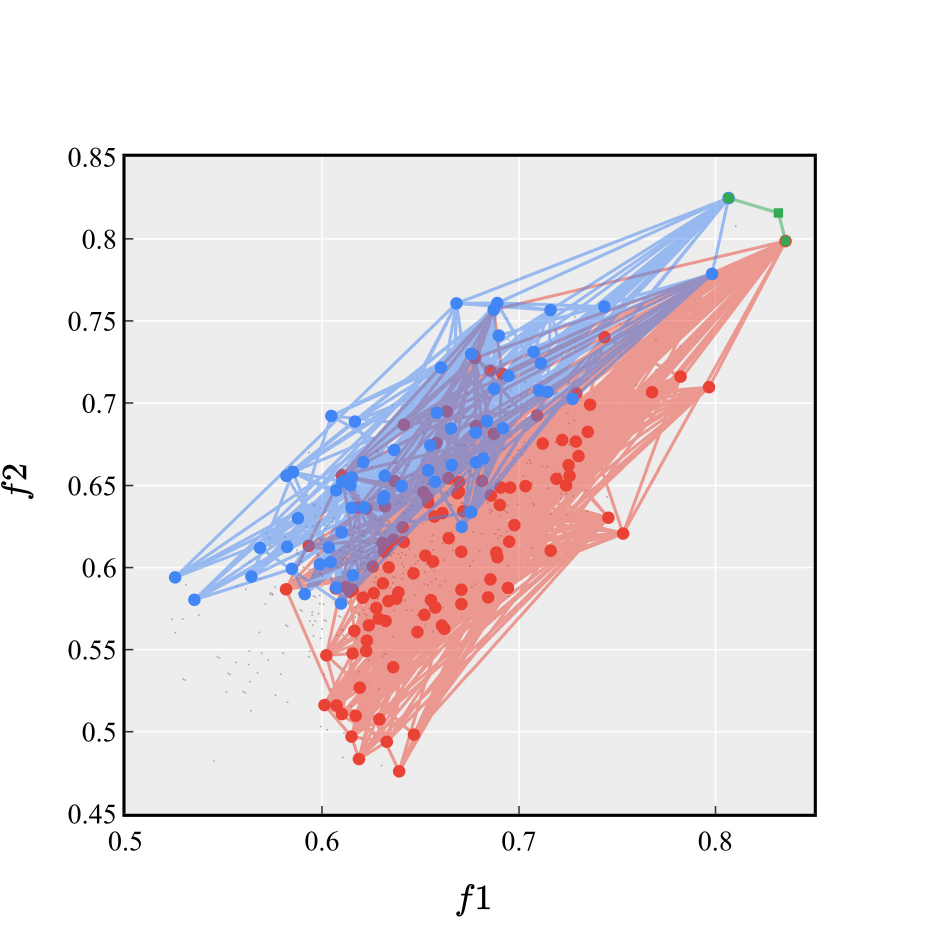}
			\\\vspace{-0.2cm}
			\subcaption{$K=3$}
		\end{center}
	\end{minipage}
	\begin{minipage}{0.245\textwidth}
		\begin{center}
			\includegraphics[width=0.95\textwidth,clip]{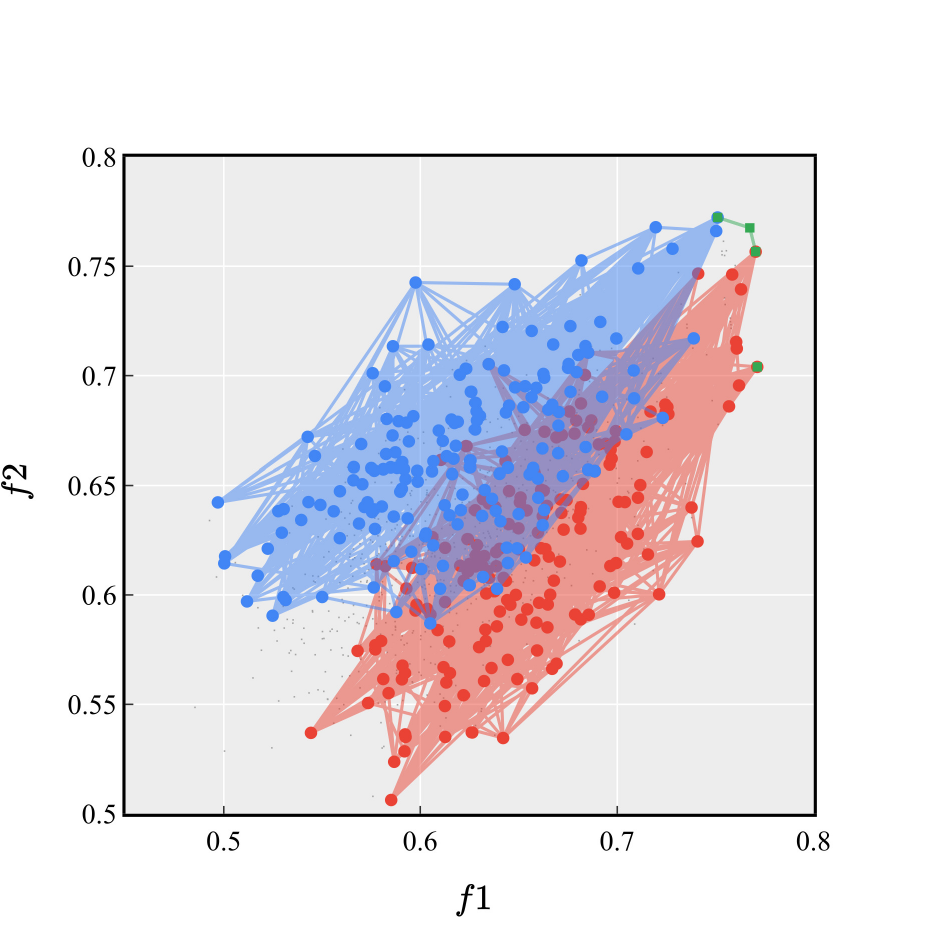}
			\\\vspace{-0.2cm}
			\subcaption{$K=4$}
		\end{center}
	\end{minipage}
\vspace{-0.1cm}
	\caption{$M=2$ objectives, correlation $\rho=0.4$, and different number of co-variables $K$}
	\label{fig:M2R0.8}
\end{figure*}

\begin{figure*}[t]
	\begin{minipage}{0.245\textwidth}
		\begin{center}
			\includegraphics[width=1.0\textwidth,clip]{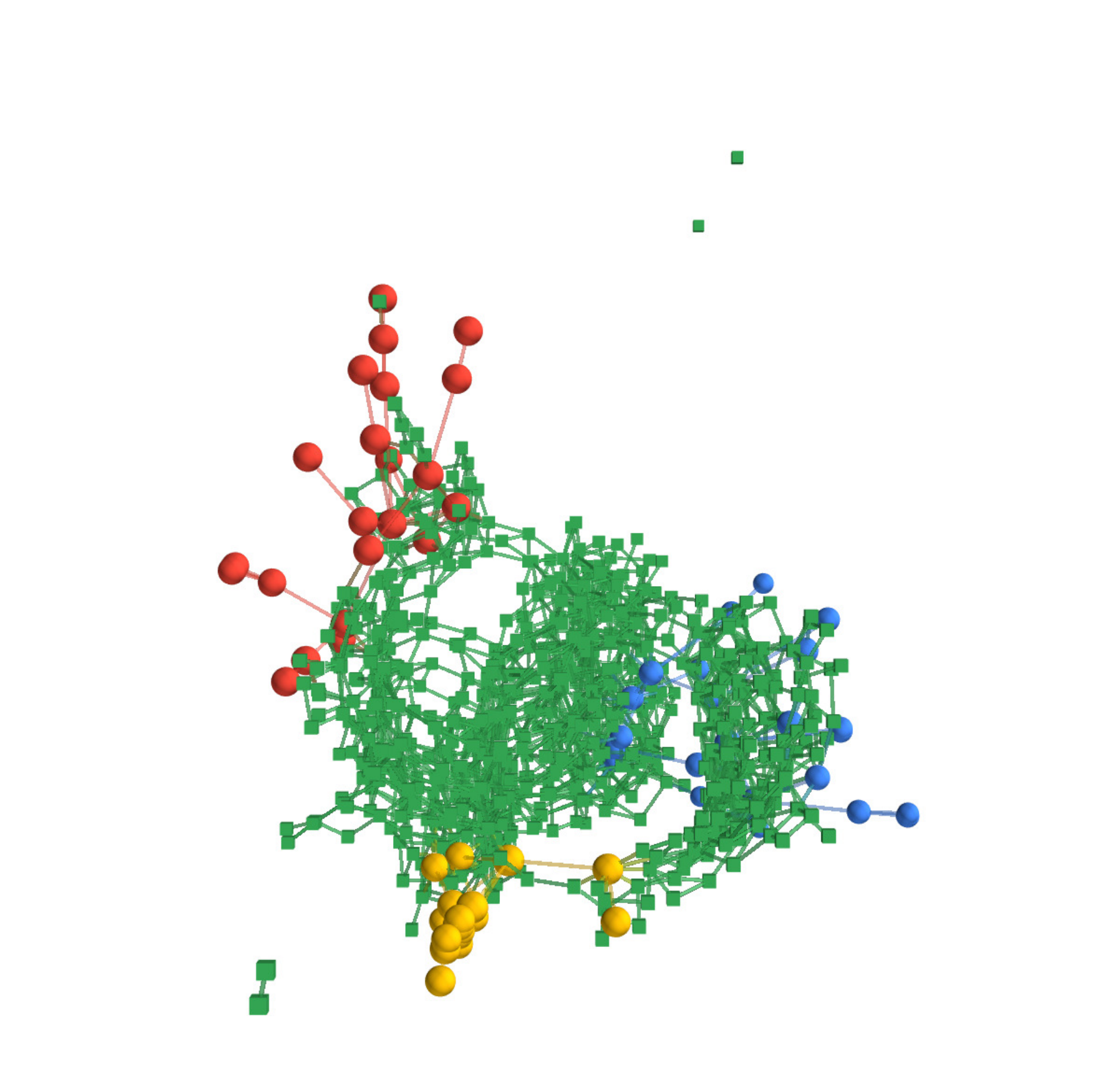}
			\\\vspace{-0.1cm}
			\subcaption{$K=1$}
		\end{center}
	\end{minipage}
	\begin{minipage}{0.245\textwidth}
		\begin{center}
			\includegraphics[width=1.0\textwidth,clip]{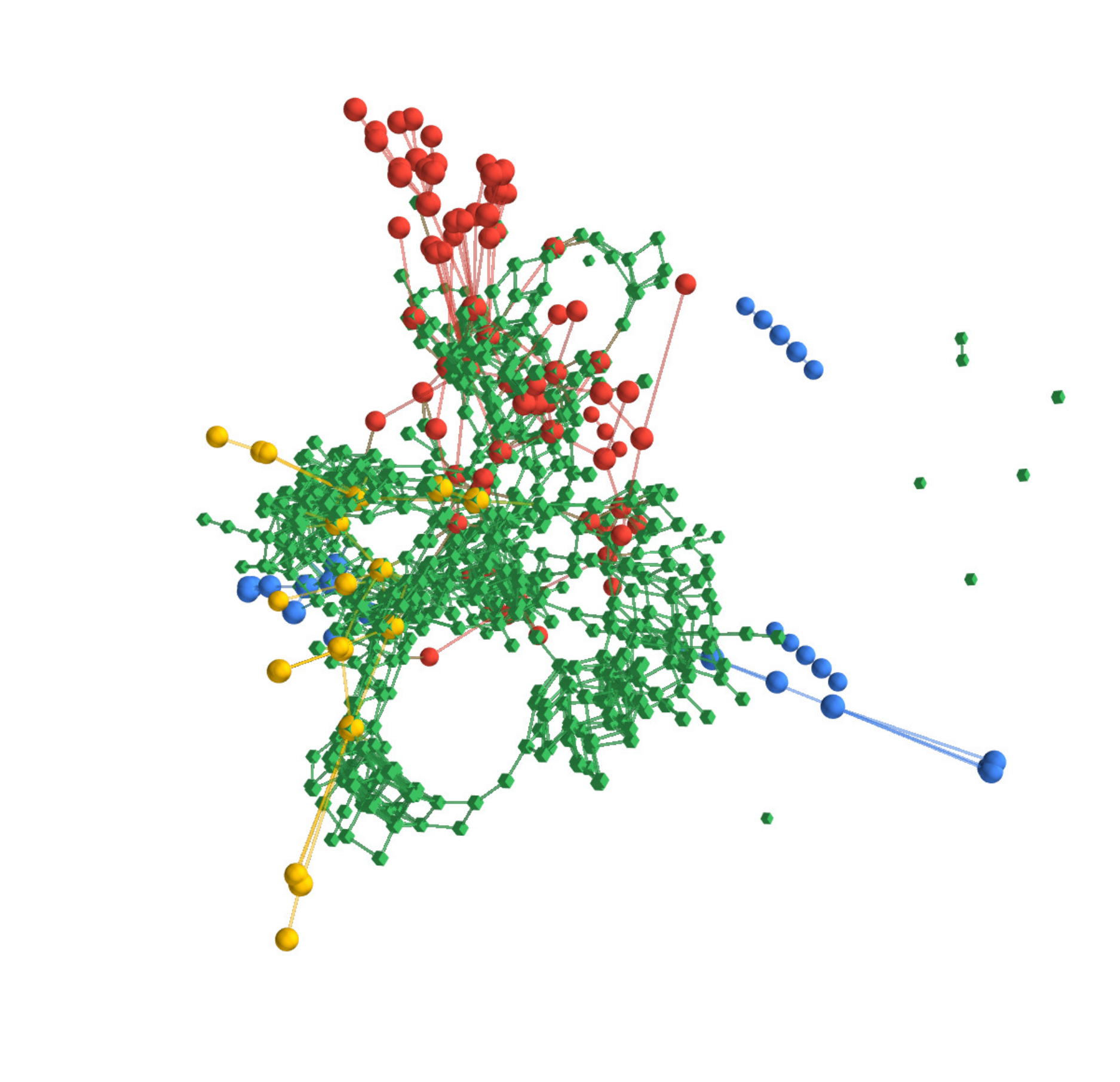}
			\\\vspace{-0.1cm}
			\subcaption{$K=2$}
		\end{center}
	\end{minipage}
	\begin{minipage}{0.245\textwidth}
		\begin{center}
			\includegraphics[width=1.0\textwidth,clip]{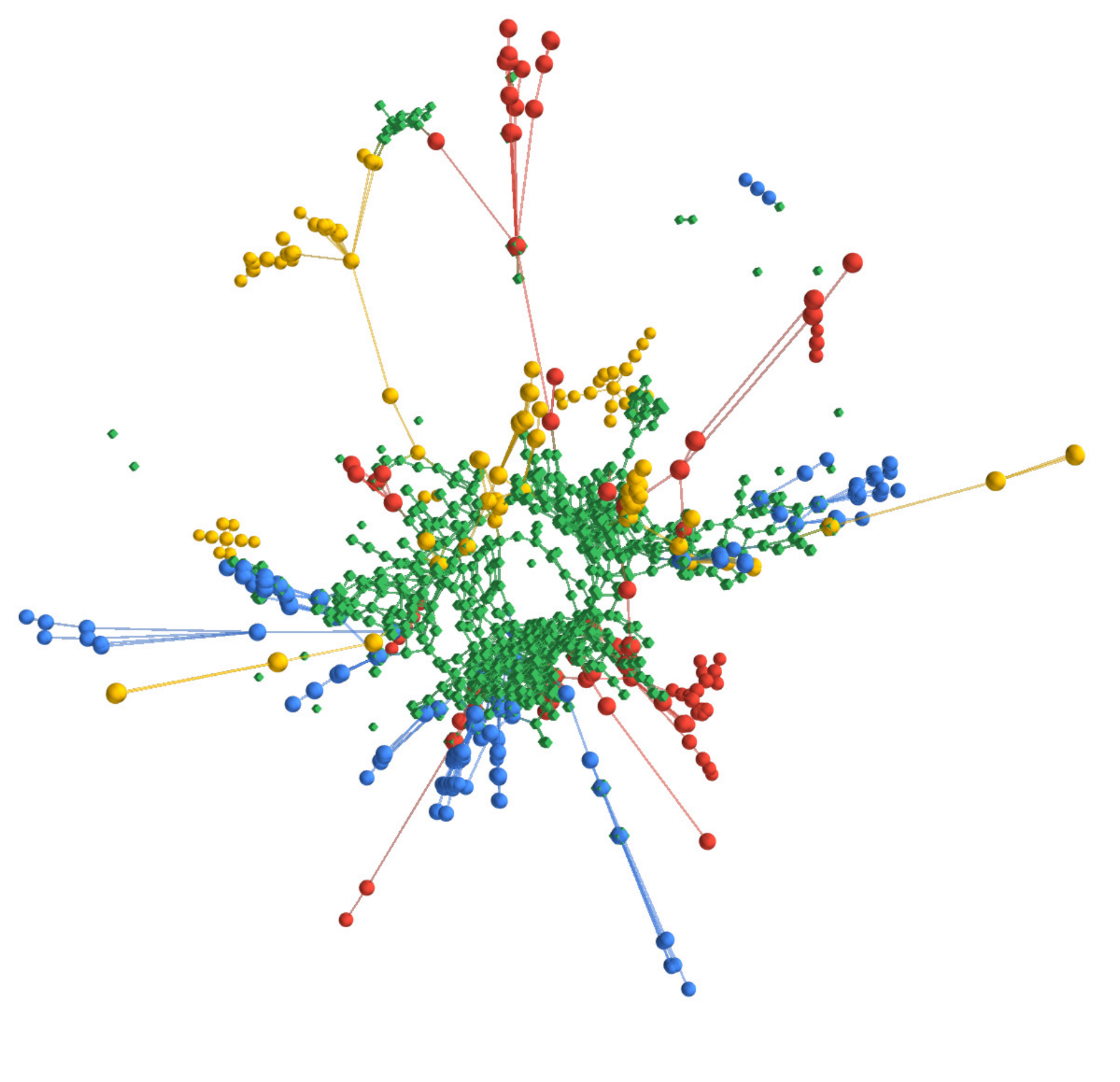}
			\\\vspace{-0.1cm}
			\subcaption{$K=3$}
		\end{center}
	\end{minipage}
	\begin{minipage}{0.245\textwidth}
		\begin{center}
			\includegraphics[width=1.0\textwidth,clip]{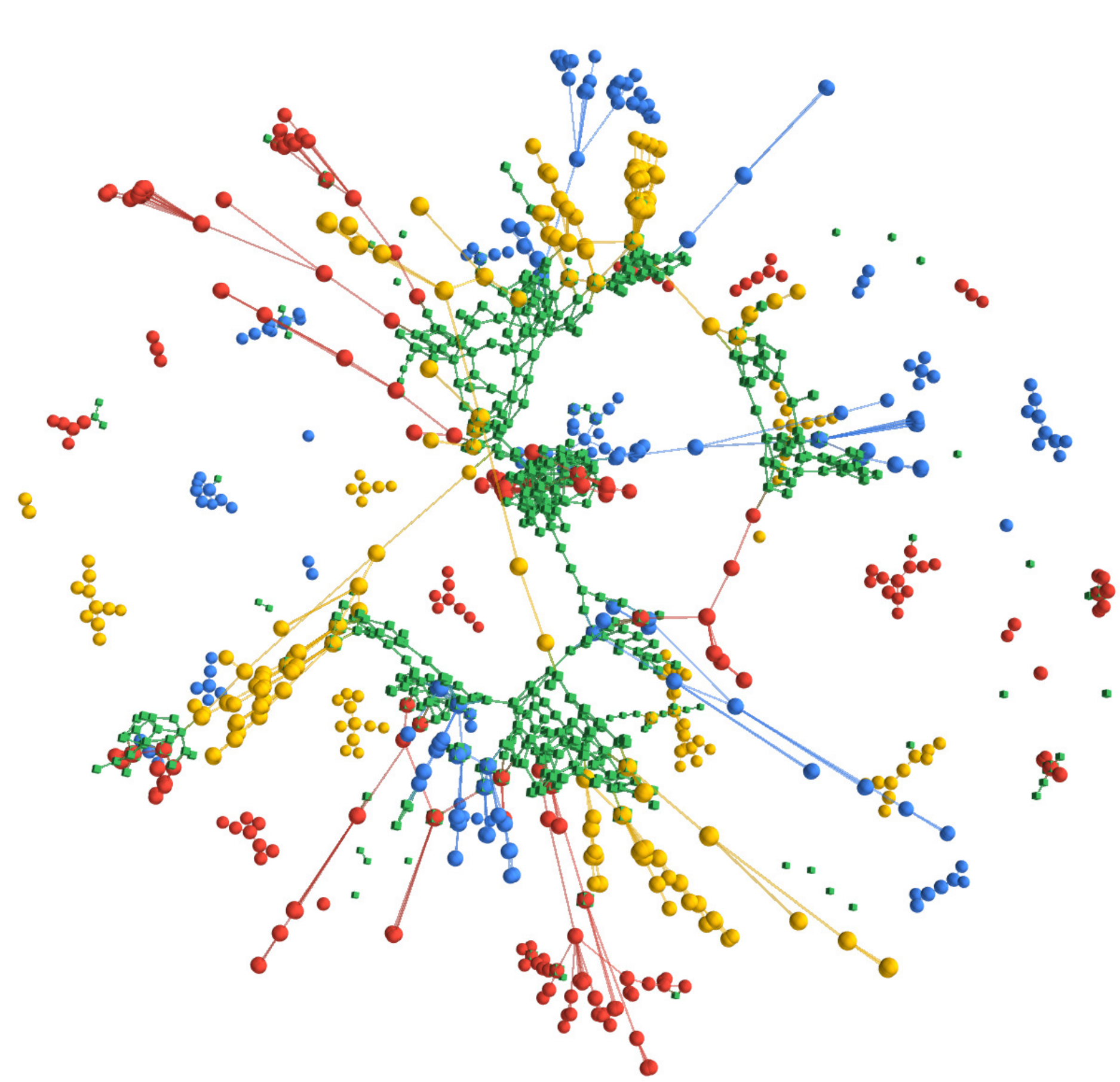}
			\\\vspace{-0.1cm}
			\subcaption{$K=4$}
		\end{center}
	\end{minipage}
	\caption{$M=3$ objectives, correlation $\rho=-0.4$, and different number of co-variables $K$}
	\label{fig:M3R-0.4}
	
	\begin{minipage}{0.245\textwidth}
		\begin{center}
			\includegraphics[width=1.0\textwidth,clip]{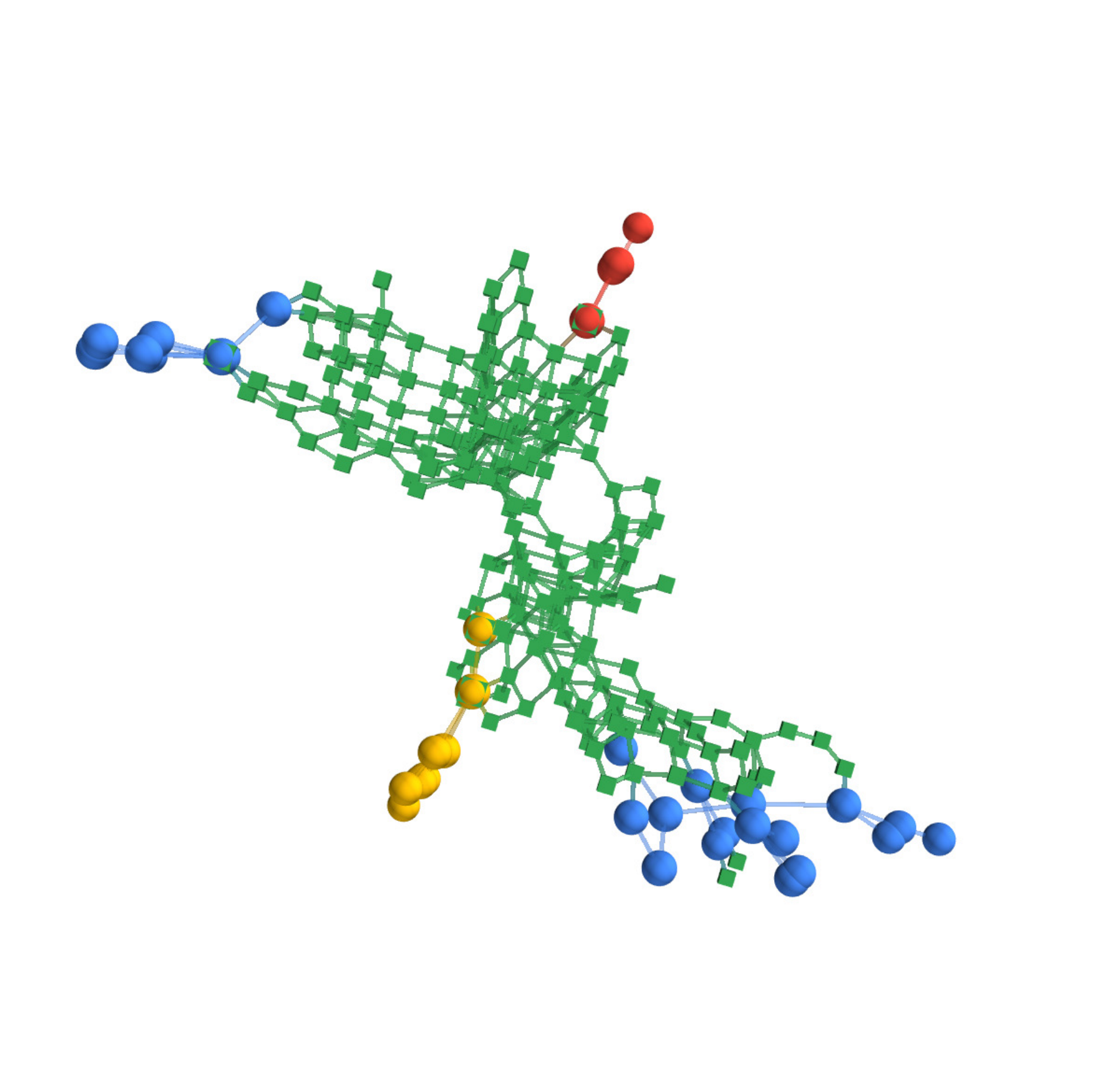}
			\\\vspace{-0.1cm}
			\subcaption{$K=1$}
		\end{center}
	\end{minipage}
	\begin{minipage}{0.245\textwidth}
		\begin{center}
			\includegraphics[width=1.0\textwidth,clip]{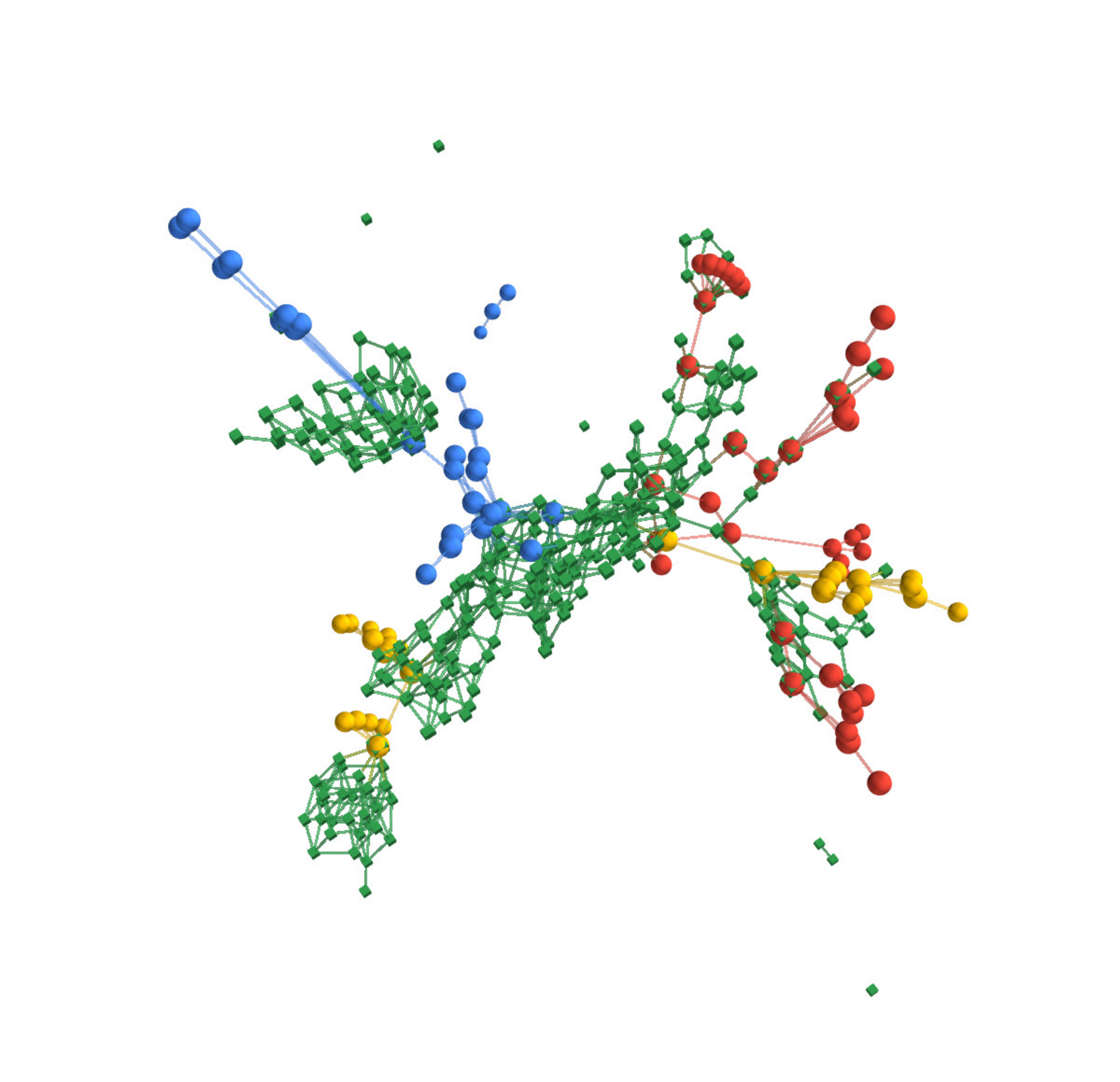}
			\\\vspace{-0.1cm}
			\subcaption{$K=2$}
		\end{center}
	\end{minipage}
	\begin{minipage}{0.245\textwidth}
		\begin{center}
			\includegraphics[width=1.0\textwidth,clip]{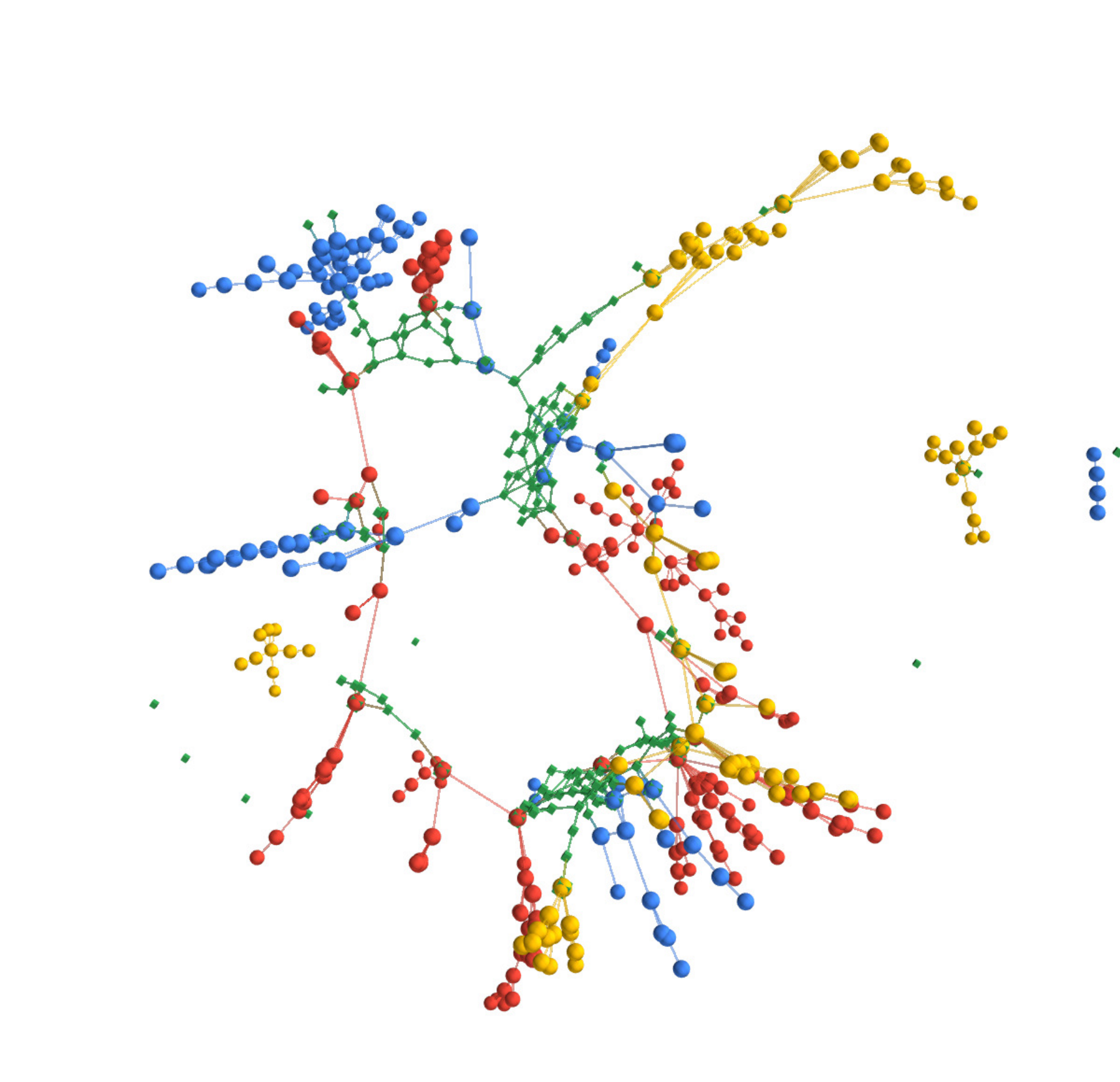}
			\\\vspace{-0.1cm}
			\subcaption{$K=3$}
		\end{center}
	\end{minipage}
	\begin{minipage}{0.245\textwidth}
		\begin{center}
			\includegraphics[width=1.0\textwidth,clip]{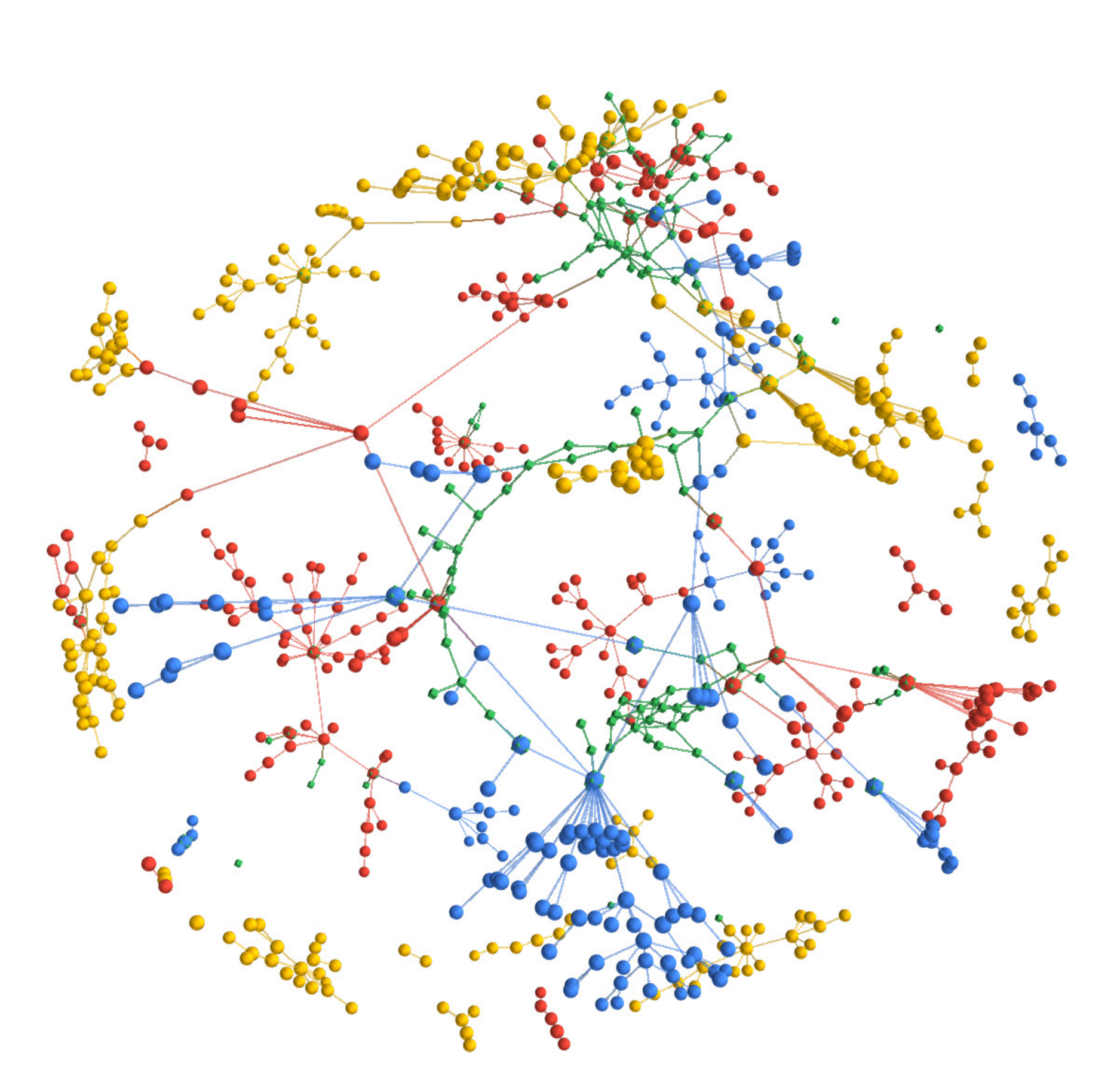}
			\\\vspace{-0.1cm}
			\subcaption{$K=4$}
		\end{center}
	\end{minipage}
	\caption{$M=3$ objectives, correlation $\rho=0.0$, and different number of co-variables $K$}
	\label{fig:M3R0.0}
	
	\begin{minipage}{0.245\textwidth}
		\begin{center}
			\includegraphics[width=1.0\textwidth,clip]{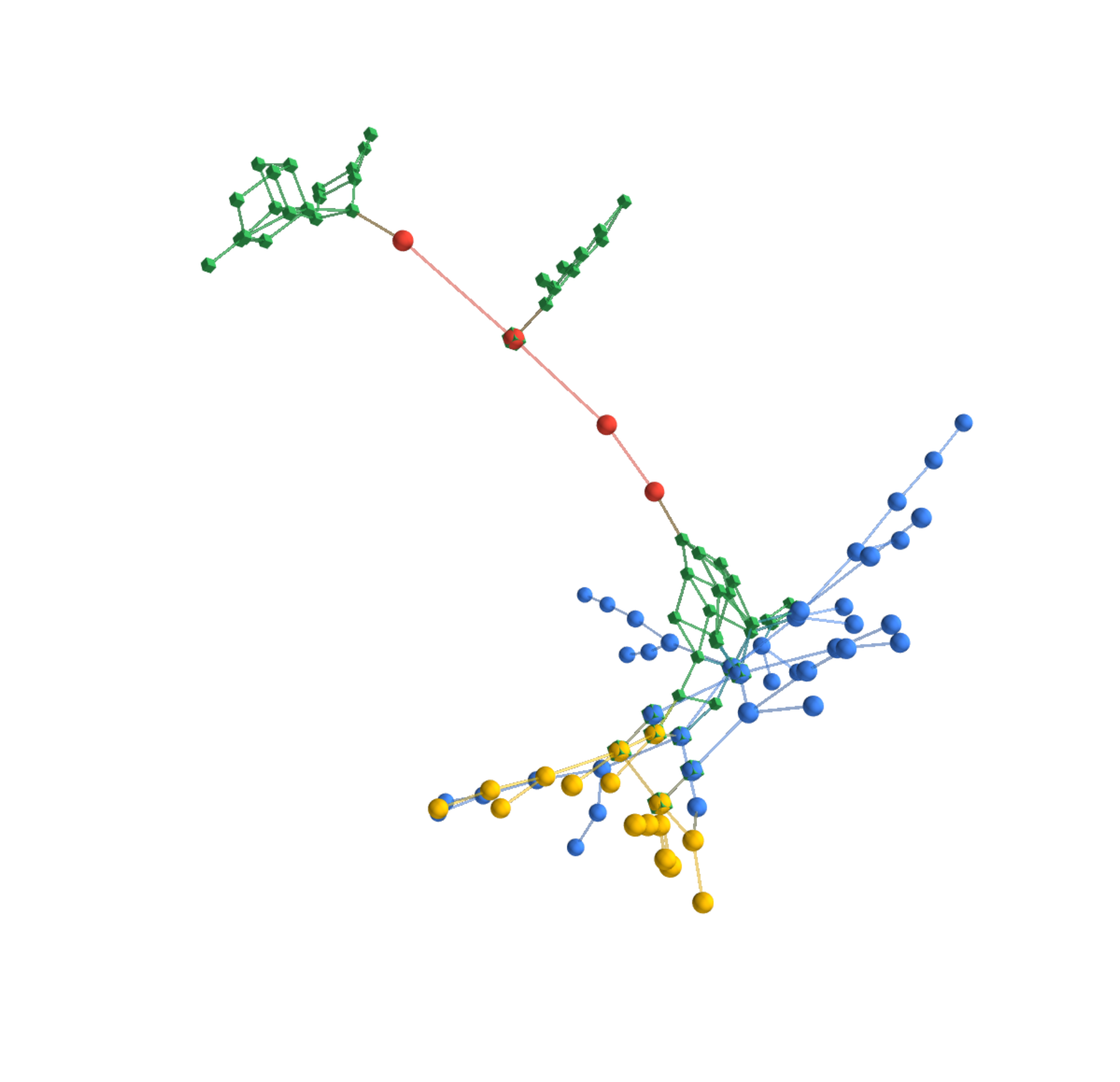}
			\\\vspace{-0.1cm}
			\subcaption{$K=1$}
		\end{center}
	\end{minipage}
	\begin{minipage}{0.245\textwidth}
		\begin{center}
			\includegraphics[width=1.0\textwidth,clip]{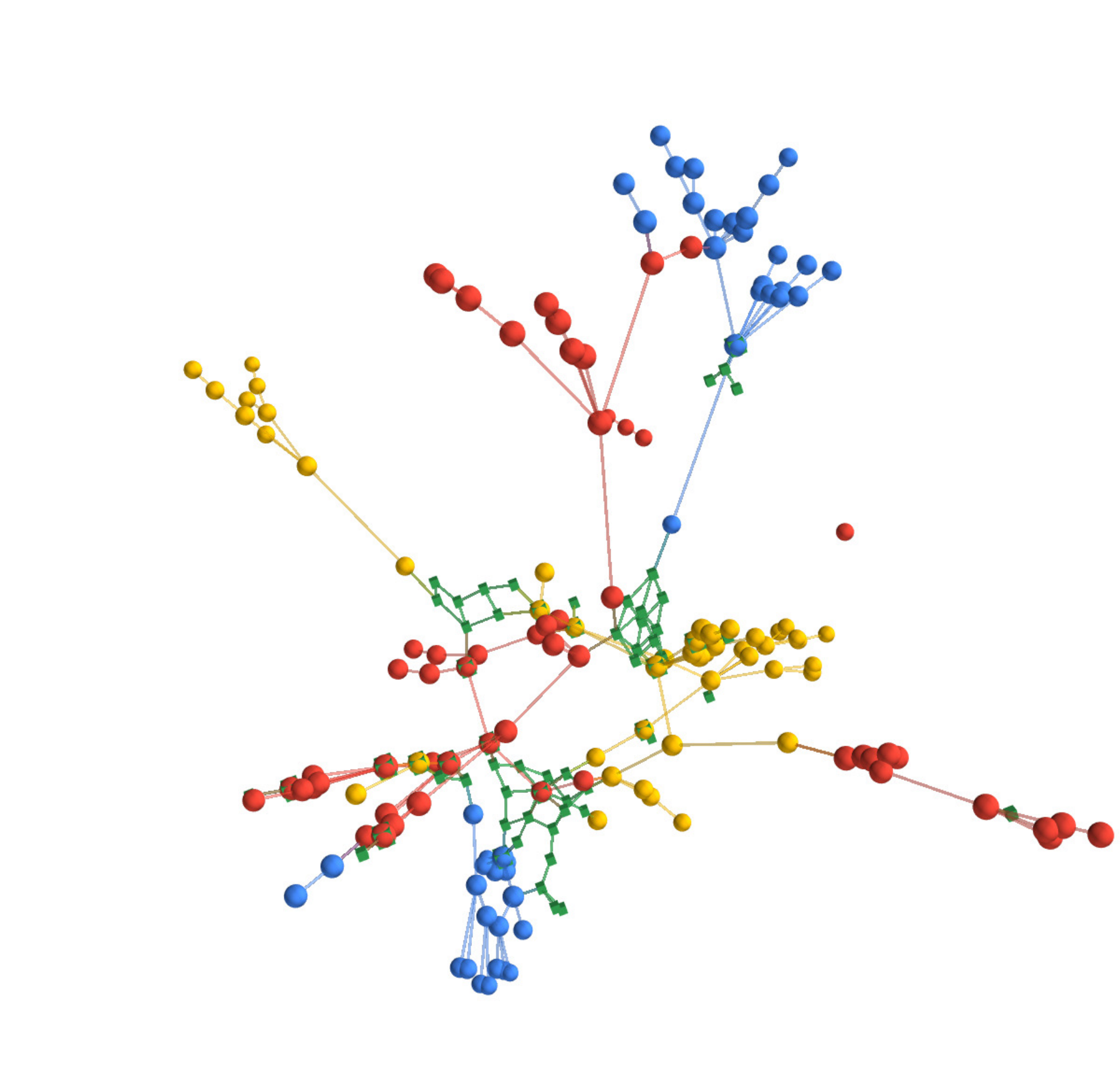}
			\\\vspace{-0.1cm}
			\subcaption{$K=2$}
		\end{center}
	\end{minipage}
	\begin{minipage}{0.245\textwidth}
		\begin{center}
			\includegraphics[width=1.0\textwidth,clip]{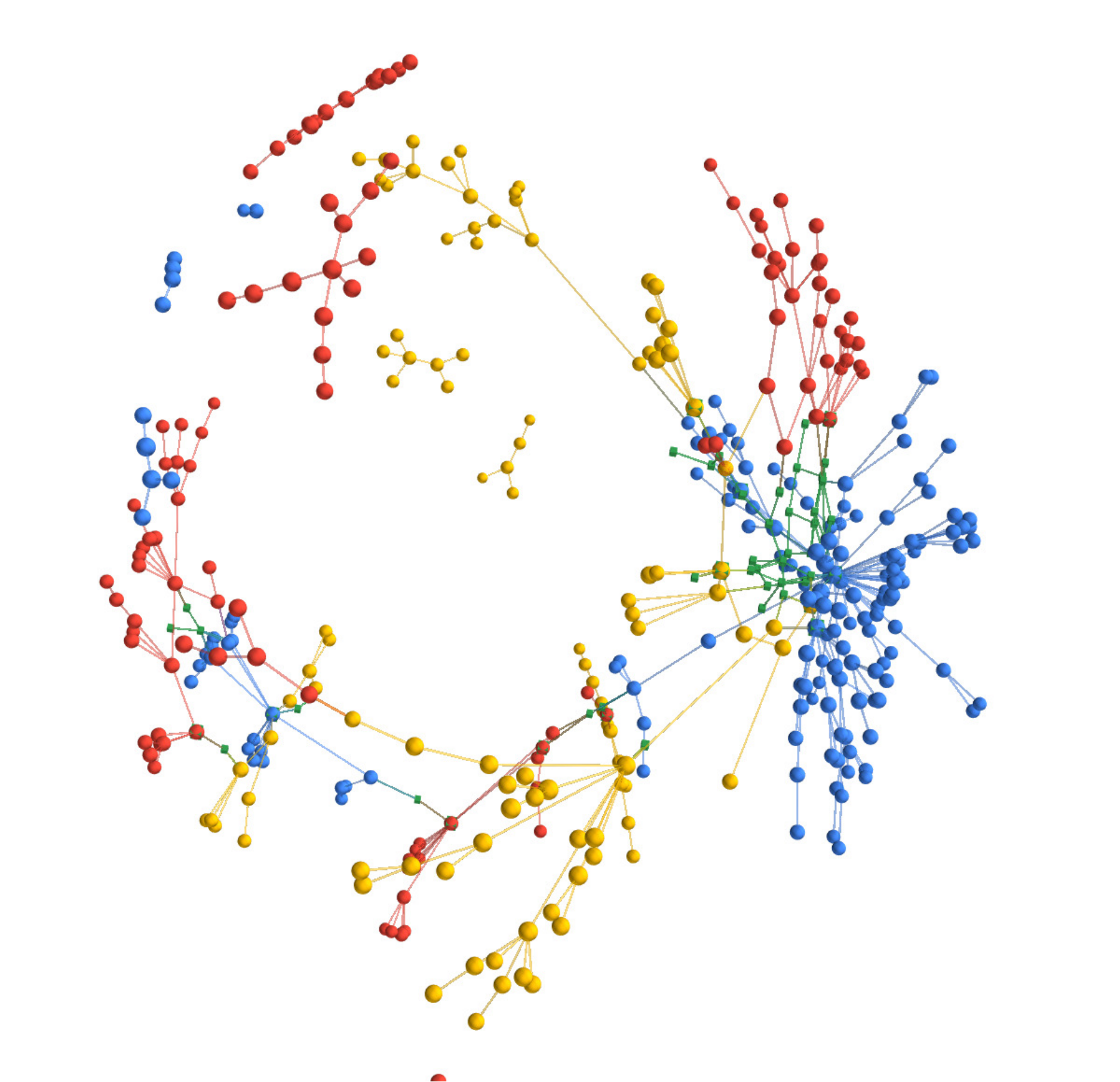}
			\\\vspace{-0.1cm}
			\subcaption{$K=3$}
		\end{center}
	\end{minipage}
	\begin{minipage}{0.245\textwidth}
		\begin{center}
			\includegraphics[width=1.0\textwidth,clip]{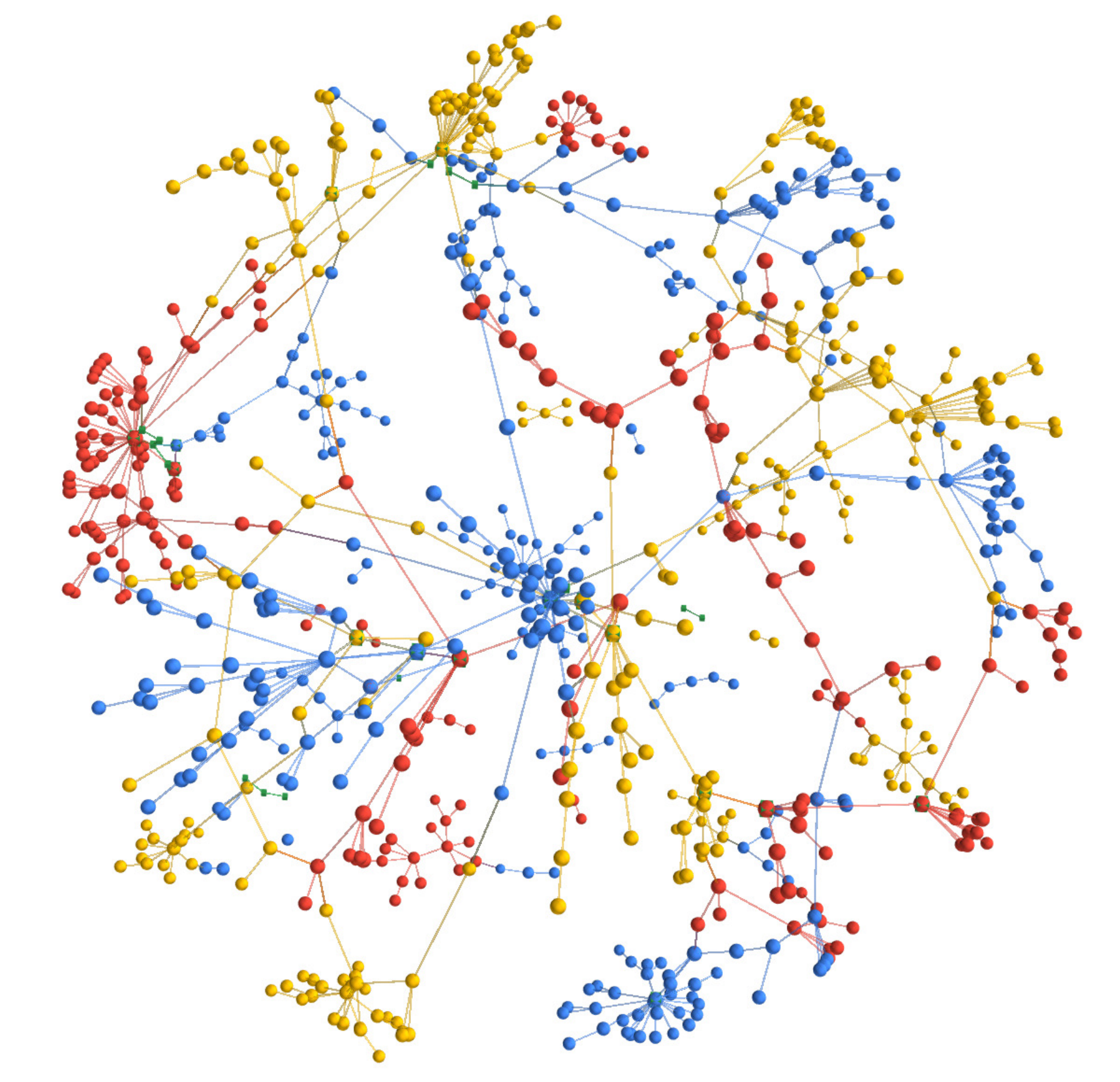}
			\\\vspace{-0.1cm}
			\subcaption{$K=4$}
		\end{center}
	\end{minipage}
	\caption{$M=3$ objectives, correlation $\rho=0.4$, and different number of co-variables $K$}
	\label{fig:M3R0.4}
\end{figure*}

\begin{figure}[t]
	\begin{center}
		\includegraphics[width=1.05\linewidth]{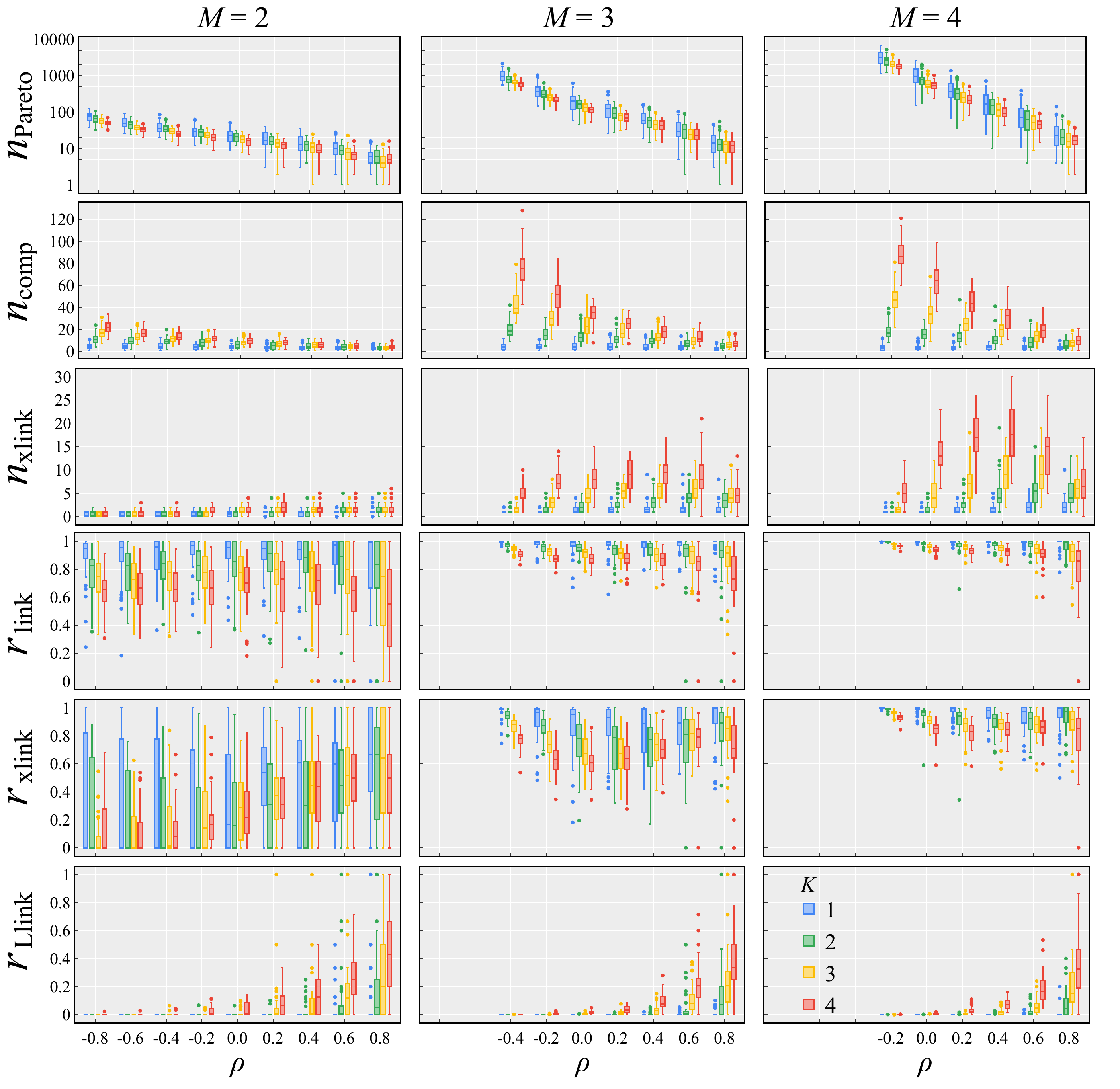}
	\end{center}
	\caption{Quantitative metrics values of the single-objective LONs and the multi-objective PON}
	\label{fig:box}
\end{figure}

\section{Qualitative Results and Discussion}
\subsection{$M=2$ Objectives}
\Figsref{fig:M2R-0.8}--\Figsrefs{fig:M2R0.8} respectively show the two single-objective LONs and the multi-objective PON on each selected problem instance with $M=2$ objectives and different numbers of co-variables $K$ and the objective correlations $\rho$. In each figure, the single-objective LONs and the multi-objective PON are plotted in $M=2$ dimensional objective space. Each maker indicates a local optimal solution in the case of the single-objective LONs or a Pareto optimal solution in the case of the multi-objective PON, and its position represents the objective values. The single-objective function $f_1$'s LON is shown in red, the single-objective function $f_2$'s LON is shown in blue, and the $M=2$ objective function vector $\bm f=(f_1,f_2)$'s PON is shown in green. Each figure also involves local Pareto optimal solution set $LP$ in gray.

We first observe the impacts of the number of co-variables $K$. From \Figsref{fig:M2R-0.8} {\bf (a)}--{\bf (d)}, we see that the number of red and blue local optimal solutions of the single-objective functions increases as the number of co-variables $K$ increases. This is a well-known problem characteristic. However, when combined with the green multi-objective PON, we see that the number of overlapped solutions between the single-objective LONs and the multi-objective PON increases as the number of co-variable $K$ increases. That is, the red and blue local optimal solutions of the single-objective functions become the green Pareto optimal solutions, especially in cases with a large $K$. We see the connectivity between the Pareto optimal solutions is high even in \Figref{fig:M2R-0.8} {\bf (d)} with a large $K$. It is suggested that searching the local optimal solutions of single-objective functions would become a way to search Pareto optimal solutions.

We next observe the impacts of the objective correlation $\rho$. From \Figsref{fig:M2R-0.8} {\bf (a)}--\Figsrefs{fig:M2R0.8} {\bf (a)}, we see that the number of green Pareto optimal solutions decreases as the objective correlation $\rho$ increases. The number of red and blue local optimal solutions also decreases. Also, we see that the objective values of the two single-objective LONs get closer as the objective correlation $\rho$ increases.

When both the objective correlation $\rho$ and the number of co-variables $K$ increase, as shown in \Figref{fig:M2R0.8} {\bf (d)}, we see the overlapped local optimal solutions between the two single-objective functions increases.

\subsection{$M=3$ Objectives}
\Figsref{fig:M3R-0.4}--\Figsrefs{fig:M3R0.4} respectively show the three single-objective LONs and the multi-objective PON on each problem with $M=3$ objectives and different numbers of co-variables $K$ and objective correlations $\rho$. Previously, \Figsref{fig:M2R-0.8}--\Figsrefs{fig:M2R0.8} in $M=2$ objective case are plotted in the objective space. Currently, \Figsref{fig:M3R-0.4}--\Figsrefs{fig:M3R0.4} in $M=3$ objective case are plotted by using the Kamada-Kawai algorithm \cite{KAMADA19897}. That is, \Figsref{fig:M3R-0.4}--\Figsrefs{fig:M3R0.4} are not in the objective space, distances between nodes are meaningless, and connected nodes are plotted closer. The single-objective function $f_1$'s LON is shown in red, the single-objective function $f_2$'s LON is shown in blue, the single-objective function $f_3$'s LON is shown in yellow, and the $M=3$ objective function vector $\bm f=(f_1,f_2,f_3)$'s PON is shown in green.

We first observe the impacts of the number of co-variables $K$. From \Figsref{fig:M3R-0.4} {\bf (a)}--{\bf (d)}, we see separating of the single-objective LONs and the multi-objective PON as the number of co-variables $K$ increases. That is, the number of funnels in each single-objective function and the number of PON components increase as the number of co-variables $K$ increases. We next observe the impacts of the objective correlation $\rho$. From \Figsref{fig:M3R-0.4} {\bf (a)}--\Figsrefs{fig:M3R0.4} {\bf (a)}, we see the number of Pareto optimal solutions decreases as the objective correlation $\rho$ increases. From \Figref{fig:M3R0.4} {\bf (a)}, we see the blue single-objective LON of the objective function $f_2$ and the yellow single-objective LON of the objective function $f_3$ are connected. Thus, funnels of single-objective functions can reach each other. It suggests if we optimize a single-objective function and obtain promising solutions, they can be utilized for an efficient optimization for other single-objective functions. When both the objective correlation $\rho$ and the number of co-variables $K$ increase, as shown in \Figref{fig:M3R0.4} {\bf (d)}, we see that the number of funnel overlaps among single-objective functions and the number of Pareto cross-links increase. In each problem, almost all parts of the green multi-objective PON have connections to the single-objective LONs. It suggests that optimizing each single-objective function can help the multi-objective optimization.

\section{Quantitative Results and Discussion}
\Figref{fig:box} shows results of quantitative metric values of the obtained single-objective LONs and the multi-objective PON on each problem. Also, \Figref{fig:correlogram} shows relations between the quantitative metric values. The difference in plot colors is the difference in the numbers of co-variables $K$. Since each problem parameter combination has fifty different instances, results are shown in the box plots and scatter plots.

\subsection{Objectives $M$}
From \Figref{fig:box}, we see that the number of Pareto optimal solutions $n_\text{Pareto}$ increases as the number of objectives $M$ increases. Also, the number of PON components $n_\text{comp}$ increases. Furthermore, we see that the number of Pareto cross-links $n_\text{xlink}$ increases as the number of objectives $M$ increases. Also, we see a tendency that the Pareto link ratio $r_\text{link}$ and the Pareto cross-link ratio $r_\text{xlink}$ increase when the number of objectives $M$ increases. In $M=4$ objective case, we see $95.6\%$ of the Pareto optimal solutions belong to the PON components of the Pareto link. That is, almost all Pareto optimal solutions are reachable from funnels of single-objective functions. Also, $95.1\%$ of Pareto links are Pareto cross-links. It reveals that the increase of the number of objectives $M$ accelerates the Pareto cross-links. In other words, almost all Pareto optimal solutions can be found by the local search from funnels of single-objective functions, especially in the case with a large number of objectives $M$.

\subsection{Objective Correlation $\rho$}
From \Figref{fig:box}, we see the number of Pareto optimal solutions $n_\text{Pareto}$ and the number of PON components $n_\text{comp}$ decrease as the objective correlation $\rho$ increase. The number of Pareto cross-links $n_\text{xlink}$ increases until the objective correlation $\rho = 0.4$ due to the approach of funnels of different single-objective functions with increasing the objective correlation $\rho$. However, the number of Pareto cross-links $n_\text{xlink}$ decreases by the further increase of the objective correlation from $\rho = 0.4$ due to the number of local optima cross-links increases instead of the Pareto cross-links when the objective correlation is high. These results reveal that the increase of the objective correlations accelerates the funnel connections by the Pareto optimal solutions. On the other hand, in $M=\{3, 4\}$ objective cases, the Pareto cross-link ratio is high even the objective correlation is negative. For the change of the objective correlation $\rho$, the change of the Pareto link ratio $r_\text{link}$ is lower than the change of the Pareto cross-link ratio $r_\text{xlink}$. These results show that the objective correlation $\rho$ impacts the Pareto cross-link ratio in the Pareto links.

\begin{figure}[t]
	\begin{center}
		\includegraphics[width=1.05\linewidth]{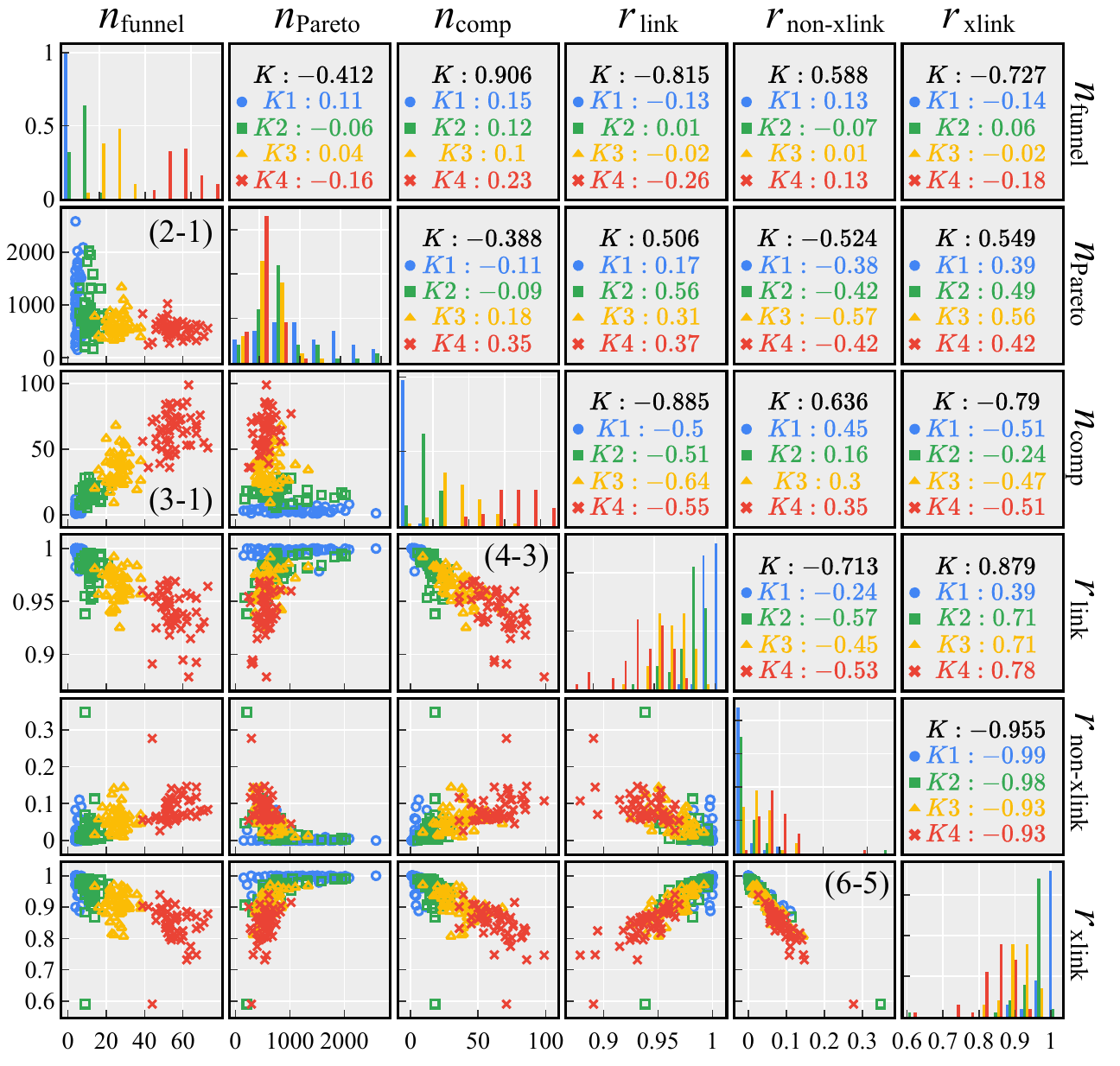}
	\end{center}
	\caption{Relations between quantitative metric values of the single-objective LONs and the multi-objective PON}
	\label{fig:correlogram}
\end{figure}

\subsection{Co-variables $K$}
The local optimal solutions and funnels increase as the number of co-variables $K$ increases. From \Figref{fig:box}, we see that the number of PON components $n_\text{comp}$ increases as the number of co-variables $K$ increases. That is, the increase of the number of co-variables $K$ accelerates the separation of the multi-objective PON. Also, we see the number of Pareto cross-links $n_\text{xlink}$ increases as the number of co-variables $K$ increases. Thus, the increase of the co-variables $K$ increases the PON components, which are many small Pareto cross-link components. From \Figref{fig:M3R-0.4} {\bf (d)} on problem with $M=3$ objectives, a negative objective correlation $\rho = -0.4$, and relatively large number of co-variables $K=4$, we see many green PON components. Also, we see these green PON components cross-link to funnels of single-objective LONs. However, funnel combinations cross-linked by each PON component are various. From these results, we see the variable diversity of the Pareto optimal solutions is high, but almost all of them are reachable from the single-objective LONs. It suggests that the optimizations of single-objective functions would accelerate multi-objective optimization involving their objective functions.

Again from \Figref{fig:box}, we see a tendency that the Pareto link ratio $r_\text{link}$ and the Pareto cross-link ratio $r_\text{xlink}$ decrease as the number of co-variables $K$ increases. They are caused by the increase of the local optima cross-link ratio $r_\text{Llink}$ with increasing the number of co-variables $K$. The transition difficulty between the Pareto optimal solutions increases when the number of co-variables increases. The number of PON components constructed by one Pareto optimal solution then increases, and the local optimal cross-link ratio $r_\text{Llink}$ is increased consequently.

\subsection{Relations Between Metrics Values}
\Figref{fig:correlogram} shows relations between the quantitative metric values of the single-objective LONs and multi-objective PONs in problems with $M=4$ objectives and the objective correlation $\rho=0$. The orthogonal cells show the distribution densities of metric values by histograms. Cells under the orthogonal cells show scatter plots based on each metric value. Cells upper the orthogonal cells show correlation coefficients of metric values. Each maker indicates one problem instance. Differences in makers' colors are differences in the numbers of co-variables $K$.

From \Figref{fig:correlogram} {\bf (2-1)} in the 2nd row of the 1st column, first we see the number of funnels $n_\text{funnel}$ and the number of Pareto optimal solutions $n_\text{Pareto}$ have no correlation. The impact of the feature of single-objective functions on the number of Pareto optimal solutions is minor. From \Figref{fig:correlogram} {\bf (3-1)} in the 3rd row of the 1st column, we see that both the number of funnels $n_\text{funnel}$ and the number of PON components $n_\text{comp}$ increase as the number of co-variables $K$ increases, and the their entire correlation coefficient becomes around 0.9.

Next, we see the number of PON components $n_\text{comp}$ and the Pareto link ratio $r_\text{link}$ are in a negative correlation, and the number of PON components $n_\text{comp}$ and the Pareto cross-link ratio $r_\text{xlink}$ are also in a negative correlation. Furthermore, from \Figref{fig:correlogram} {\bf (6-5)} in the 6th row of the 5th column, we see the $r_\text{non-xlink}$ and $r_\text{xlink}$ are in a highly negative correlation.

\subsection{Discussion Summary}
We showed that most Pareto optimal solutions were reachable from the single-objective local optimal solutions in all problems used in this work. It suggests that the search for single-objective functions assists multi-objective optimization. Other detailed findings are as follows.

The number of objectives $M$ impacts the Pareto link ratio and the Pareto cross-link ratio. Most Pareto optimal solutions belong to PON components cross-linking multiple single-objective LONs when the number of objectives $M$ increases. The increase of the objective correlation $\rho$ makes funnels of single-objective functions closer and decreases the number of Pareto optimal solutions and the number of PON components. However, the increase of the objective correlation $\rho$ increases the number of Pareto cross-links. A high objective correlation $\rho$ brings local optima cross-links instead of the Pareto optima cross-links. The number of co-variables $K$ impacts the number of Pareto cross-links. A high $K$ arises Pareto cross-links between multiple funnels and increases the Pareto links not cross-linking.

\section{Conclusions}
To reveal the impacts of single-objective landscapes of multi-objective optimization problems on the distribution of the Pareto optimal solutions and their findability, we analyzed the relations between the single-objective LONs and the multi-objective PON by using $\rho MNK$ landscape problems. Experimental results showed that most Pareto optimal solutions could be reached from single-objective local optimal solutions. This tendency was emphasized by the increase of the number of objectives $M$ and the objective correlation $\rho$. Also, we showed that the rise of the number of co-variables $K$ impacted the number of Pareto cross-links.

As future works, we will study the multi-objective optimizer efficiently searching the Pareto optimal solutions using multiple single-objective optimizations and Pareto cross-links.

\bibliographystyle{IEEEtran}
\bibliography{reference}

\begin{thebibliography}{10}
\providecommand{\url}[1]{#1}
\csname url@samestyle\endcsname
\providecommand{\newblock}{\relax}
\providecommand{\bibinfo}[2]{#2}
\providecommand{\BIBentrySTDinterwordspacing}{\spaceskip=0pt\relax}
\providecommand{\BIBentryALTinterwordstretchfactor}{4}
\providecommand{\BIBentryALTinterwordspacing}{\spaceskip=\fontdimen2\font plus
\BIBentryALTinterwordstretchfactor\fontdimen3\font minus
  \fontdimen4\font\relax}
\providecommand{\BIBforeignlanguage}[2]{{%
\expandafter\ifx\csname l@#1\endcsname\relax
\typeout{** WARNING: IEEEtran.bst: No hyphenation pattern has been}%
\typeout{** loaded for the language `#1'. Using the pattern for}%
\typeout{** the default language instead.}%
\else
\language=\csname l@#1\endcsname
\fi
#2}}
\providecommand{\BIBdecl}{\relax}
\BIBdecl

\bibitem{coello2007}
C.~A.~C. Coello, G.~B. Lamont, D.~A. Van~Veldhuizen \emph{et~al.},
  \emph{Evolutionary algorithms for solving multi-objective problems}.\hskip
  1em plus 0.5em minus 0.4em\relax Springer, 2007, vol.~5.

\bibitem{tuvsar2014}
T.~Tu{\v{s}}ar and B.~Filipi{\v{c}}, ``Visualization of pareto front
  approximations in evolutionary multiobjective optimization: A critical review
  and the prosection method,'' \emph{IEEE Transactions on Evolutionary
  Computation}, vol.~19, no.~2, pp. 225--245, 2014.

\bibitem{ochoa2008}
G.~Ochoa, M.~Tomassini, S.~V{\'e}rel, and C.~Darabos, ``A study of nk
  landscapes' basins and local optima networks,'' in \emph{Proceedings of the
  10th annual conference on Genetic and evolutionary computation}, 2008, pp.
  555--562.

\bibitem{paquete2009}
L.~Paquete and T.~St{\"u}tzle, ``Clusters of non-dominated solutions in
  multiobjective combinatorial optimization: An experimental analysis,'' in
  \emph{Multiobjective Programming and Goal Programming}.\hskip 1em plus 0.5em
  minus 0.4em\relax Springer, 2009, pp. 69--77.

\bibitem{liefooghe2018}
A.~Liefooghe, B.~Derbel, S.~Verel, M.~L{\'o}pez-Ib{\'a}{\~n}ez, H.~Aguirre, and
  K.~Tanaka, ``On pareto local optimal solutions networks,'' in
  \emph{International Conference on Parallel Problem Solving from
  Nature}.\hskip 1em plus 0.5em minus 0.4em\relax Springer, 2018, pp. 232--244.

\bibitem{liefooghe2019}
A.~Liefooghe, F.~Daolio, S.~Verel, B.~Derbel, H.~Aguirre, and K.~Tanaka,
  ``Landscape-aware performance prediction for evolutionary multiobjective
  optimization,'' \emph{IEEE Transactions on Evolutionary Computation},
  vol.~24, no.~6, pp. 1063--1077, 2019.

\bibitem{verel2011_2}
S.~Verel, A.~Liefooghe, L.~Jourdan, and C.~Dhaenens, ``Analyzing the effect of
  objective correlation on the efficient set of mnk-landscapes,'' in
  \emph{International Conference on Learning and Intelligent
  Optimization}.\hskip 1em plus 0.5em minus 0.4em\relax Springer, 2011, pp.
  116--130.

\bibitem{kauffman1987}
S.~Kauffman and S.~Levin, ``Towards a general theory of adaptive walks on
  rugged landscapes,'' \emph{Journal of theoretical Biology}, vol. 128, no.~1,
  pp. 11--45, 1987.

\bibitem{aguirre2004}
H.~E. Aguirre and K.~Tanaka, ``Insights on properties of multiobjective
  mnk-landscapes,'' in \emph{Proceedings of the 2004 Congress on Evolutionary
  Computation (IEEE Cat. No. 04TH8753)}, vol.~1.\hskip 1em plus 0.5em minus
  0.4em\relax IEEE, 2004, pp. 196--203.

\bibitem{allmendinger2022}
R.~Allmendinger, A.~Jaszkiewicz, A.~Liefooghe, and C.~Tammer, ``What if we
  increase the number of objectives? theoretical and empirical implications for
  many-objective combinatorial optimization,'' \emph{Computers \& Operations
  Research}, p. 105857, 2022.

\bibitem{ochoa2017}
G.~Ochoa, N.~Veerapen, F.~Daolio, and M.~Tomassini, ``Understanding phase
  transitions with local optima networks: number partitioning as a case
  study,'' in \emph{European Conference on Evolutionary Computation in
  Combinatorial Optimization}.\hskip 1em plus 0.5em minus 0.4em\relax Springer,
  2017, pp. 233--248.

\bibitem{KAMADA19897}
T.~Kamada and S.~Kawai, ``An algorithm for drawing general undirected graphs,''
  \emph{Information Processing Letters}, vol.~31, no.~1, pp. 7--15, 1989.

\end{thebibliography}
\balance
\end{document}